\documentclass[acmtog]{acmart}
\setcopyright{none}
\renewcommand\footnotetextcopyrightpermission[1]{}
\acmDOI{}
\acmISBN{}
\acmConference{}{}{}
\acmYear{2026}
\copyrightyear{2026}

\acmJournal{TOG}
\usepackage{booktabs}
\usepackage{tabularx}
\usepackage{makecell}
\usepackage{multirow}
\usepackage[table]{xcolor}
\usepackage{array}
\usepackage{amsmath}

\usepackage{pifont}

\usepackage{enumitem}

\newcolumntype{Y}{>{\centering\arraybackslash}X}
\newcolumntype{V}{!{\hspace{6pt}\vrule width 0.6pt\hspace{6pt}}}

\definecolor{tabhighlight}{RGB}{245,245,245}

\definecolor{metricbench}{RGB}{60,120,200}
\definecolor{metricreal}{RGB}{200,120,60}
\definecolor{metricavg}{RGB}{90,90,90}

\newcommand{\cmark}{\ding{51}} % ✓

\newcommand{\method}{{$\mathbf{GIVE}$}} % Geometric Instruction Video Editing
\newcommand{\dataset}{$\mathbf{GIVE}$-Data}
\newcommand{\benchmark}{$\mathbf{GIVE}$-Bench}

\begin{document}
\title{Geometry-Instructed Video Editing}
\author{Chirui Chang}
\affiliation{%
  \institution{The University of Hong Kong}
  \city{Hong Kong}
  \country{Hong Kong}}

\author{Xiaoyang Lyu}
\affiliation{%
  \institution{The University of Hong Kong}
  \city{Hong Kong}
  \country{Hong Kong}}

\author{Yi-Hua Huang}
\affiliation{%
  \institution{The University of Hong Kong}
  \city{Hong Kong}
  \country{Hong Kong}}

\author{Haoru Tan}
\affiliation{%
  \institution{The University of Hong Kong}
  \city{Hong Kong}
  \country{Hong Kong}}

\author{Shizhen Zhao}
\affiliation{%
  \institution{The University of Hong Kong}
  \city{Hong Kong}
  \country{Hong Kong}}

\author{Yikang Ding}
\affiliation{%
  \institution{Kuaishou Technology}
  \country{China}}

\author{Jianmin Bao}
\affiliation{%
  \institution{Kuaishou Technology}
  \country{China}}

\author{Xin Tao}
\affiliation{%
  \institution{Kuaishou Technology}
  \country{China}}

\author{Pengfei Wan}
\affiliation{%
  \institution{Kuaishou Technology}
  \country{China}}

\author{Xiaojuan Qi}
\affiliation{%
  \institution{The University of Hong Kong}
  \city{Hong Kong}
  \country{Hong Kong}}

\renewcommand{\shortauthors}{Chang et al.}
\authorsaddresses{}

\begin{abstract}
Object-level geometric edits, including translating, rotating, scaling, duplicating, or removing an object, are routine operations in digital content creation (DCC) workflows, yet they remain unreliable in generative video editing. 
The key challenge lies in specifying the target object's 3D state change unambiguously across viewpoint and time, while consistently updating geometry-dependent secondary effects such as shadows and reflections.
We introduce {\method}, a geometry-instructed video editing framework that represents edits through a unified object-state formulation. 
Two video-aligned geometry streams describe the target object before and after editing: a \emph{depth-box} encoding coarse 3D placement and extent, and an \emph{orientation-box} providing an appearance-agnostic orientation cue. 
Together, these streams provide a compact pre/post geometric specification for object-state transitions. 
To provide paired supervision for learning these edits, we build a scalable graphics-engine pipeline that executes object-level edit programs and renders controlled before/after pairs, isolating the intended geometric edit while keeping secondary effects consistent with the transformation.
Experimental results demonstrate that \method\ produces faithful geometric edits with temporal coherence and consistent secondary effects across operators in a unified framework, and shows promising transfer to in-the-wild videos.
\end{abstract}

\begin{CCSXML}
<ccs2012>
   <concept>
       <concept_id>10010147.10010178.10010224</concept_id>
       <concept_desc>Computing methodologies~Computer vision</concept_desc>
       <concept_significance>500</concept_significance>
       </concept>
 </ccs2012>
\end{CCSXML}

\ccsdesc[500]{Computing methodologies~Computer vision}


% \keywords{Do, Not, Use, This, Code, Put, the, Correct, Terms, for,
\keywords{3D-aware video editing, Synthetic data, Generative models}
\begin{teaserfigure}
  \centering
  \includegraphics[width=0.95\textwidth]{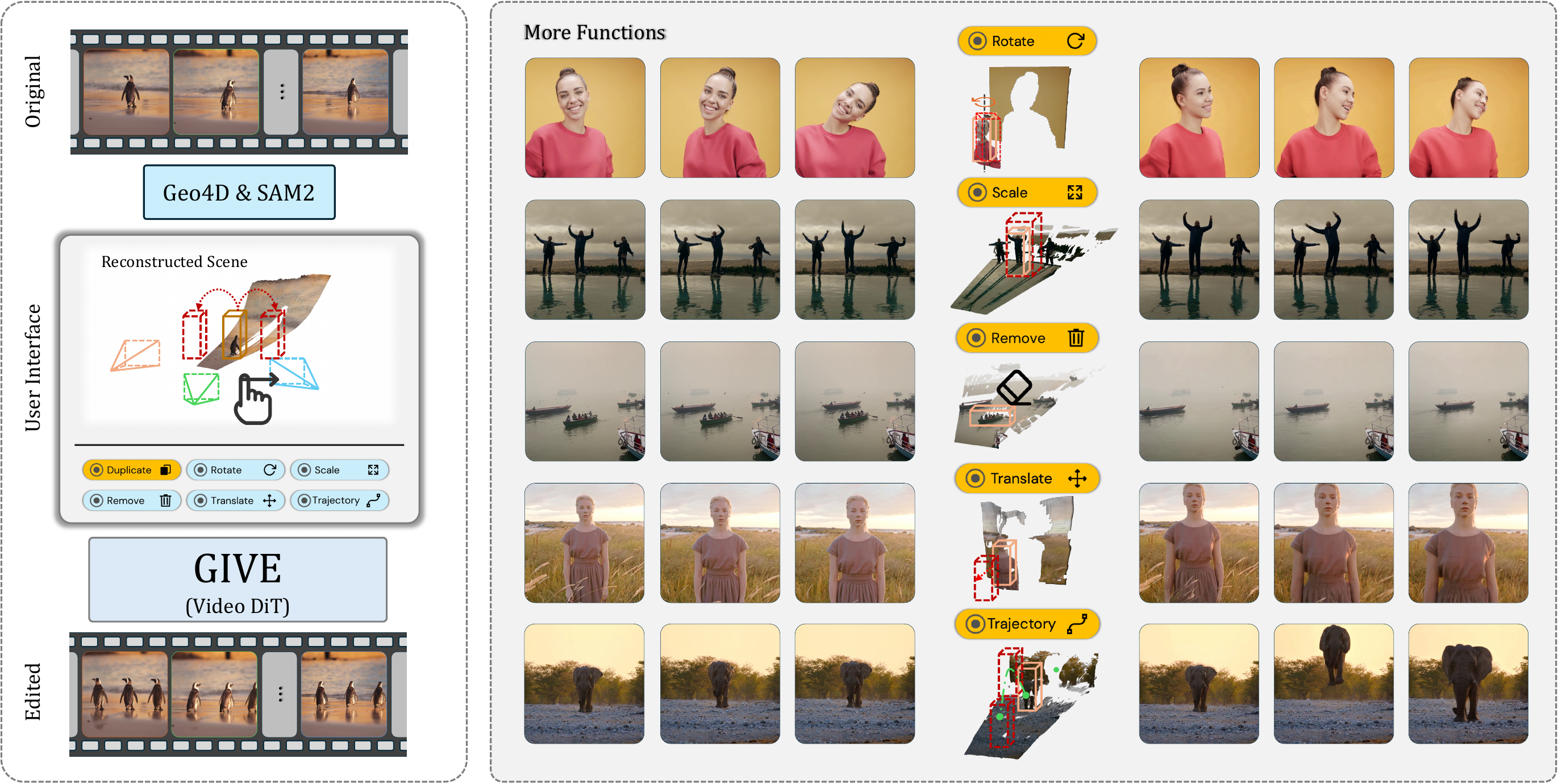}
  \caption{\method\ performs object-level geometric video editing using unified pre and post box-based geometry instructions derived from lightweight user interaction and off-the-shelf tools. A single Video DiT editor supports multiple DCC-style operators, including rotation, scaling, removal, translation, duplication, and trajectory editing, while maintaining temporal coherence and producing consistent geometry-dependent secondary effects.}
  \label{fig:teaser}
\end{teaserfigure}

\maketitle

\section{Introduction}
Generative video models are increasingly being explored as tools for digital content creation (DCC), where artists and designers need not only to synthesize new footage but also to iteratively revise existing content~\cite{openai2025sora2,deepmind2025veo3,deepmind2025genie3,kuaishou2025kling,minimax2025hailuo,ByteDance2026Seedance}. 
In practical DCC workflows, many revisions are object-level geometric operations: nudging a prop to refine composition, rotating an actor to face the camera, scaling an object for emphasis, or removing a distraction. 
While such edits are routine in traditional graphics pipelines, they remain unreliable in current generative video editors, especially when the intended 3D change must stay precise and consistent across viewpoint and time.

\begin{figure}[t]
  \centering
  \includegraphics[width=\columnwidth]{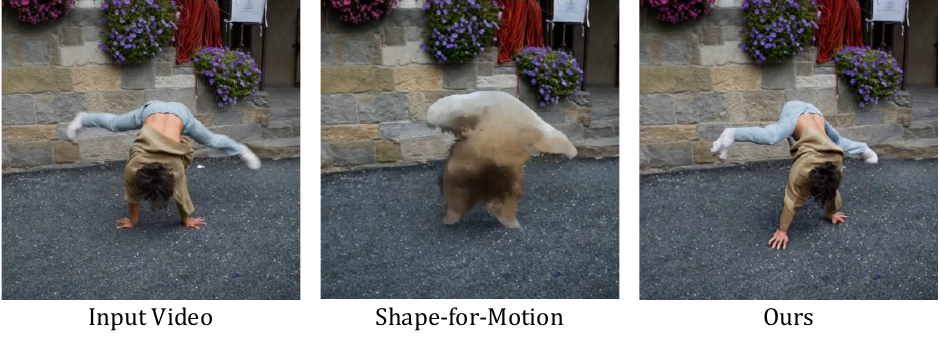}
  \vspace{-25pt}
  \caption{\textbf{Limitation of reconstruction-based methods.}
Shape-for-Motion fails to recover a coherent proxy with complex motion, leading to unreliable edits; \method\ remains stable without per-video reconstruction/optimization.}
  \label{fig:intro_fail}
  \vspace{-15pt}
\end{figure}

\begin{figure*}[p]
  \centering
  \includegraphics[width=\textwidth,height=0.95\textheight,keepaspectratio]{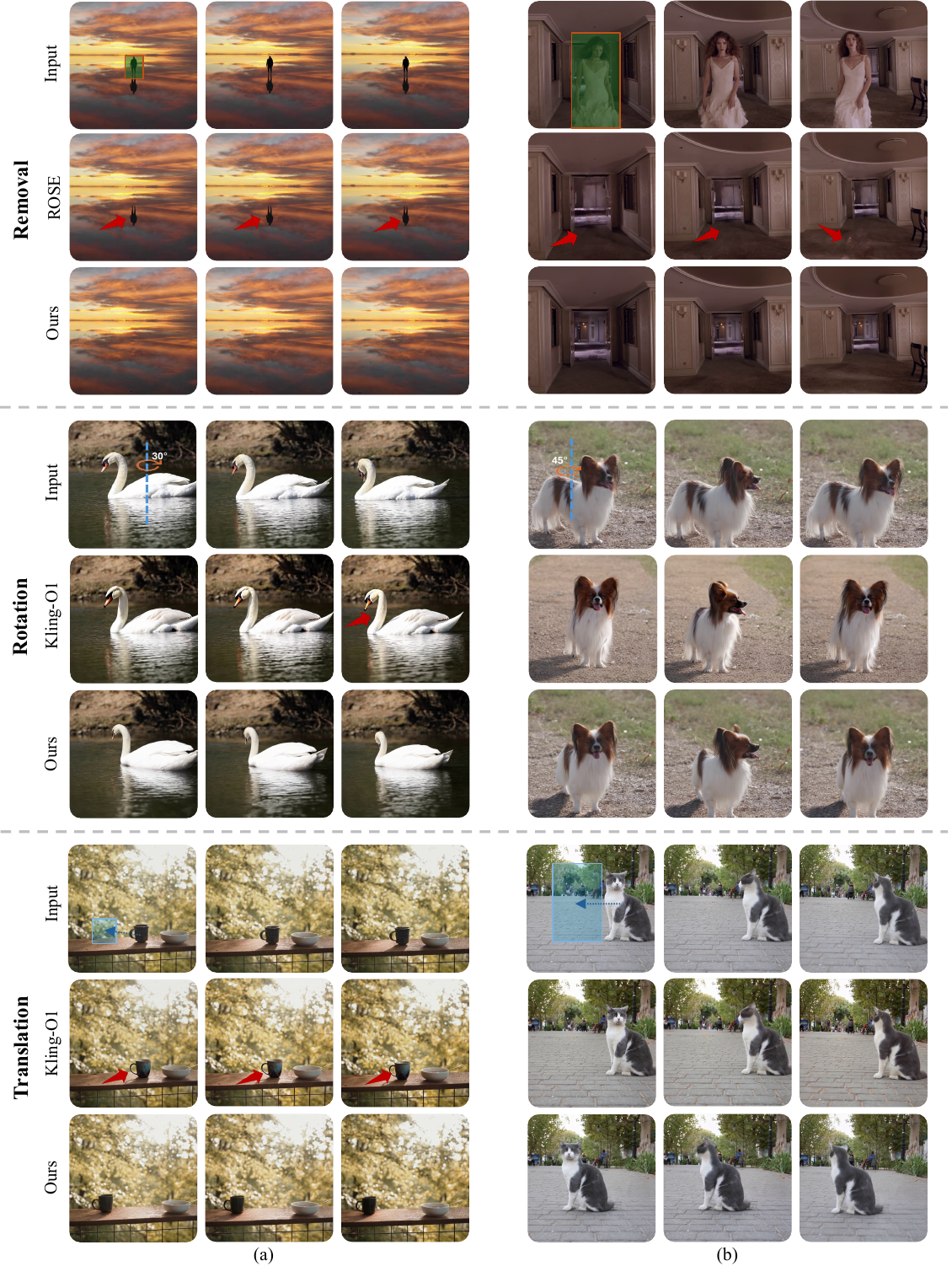}
  \vspace{-6pt}
  \caption{\textbf{Qualitative comparison on real videos (I).}
We show results for geometric operators including removal, rotation and translation on the real videos.
% For each example, we demonstrate the input and instruction, followed by edited outputs from \method\ and the baselines under the same setting.
Across cases, \method\ better follows the intended geometric manipulation while preserving object identity and non-edited regions, and produces more coherent geometry-dependent secondary effects (e.g., shadows and reflections).
Baselines often exhibit operator mis-execution, inconsistent secondary-effect updates, and identity/background drift, as highlighted in the annotated regions.
}
  \label{fig:exp1}
\end{figure*}

Existing work addresses object-level video revision through \emph{image-space} or semantic controls, such as text prompts, masks, or tracks~\cite{team2025kling,miao2025rose,gu2025diffusion}.
These interfaces are convenient and often produce visually compelling results, but they do not explicitly determine the target object's pre- and post-edit 3D state.
As a result, the intended transformation is only weakly constrained: the model may under-execute or mis-execute the operation, and geometry-dependent effects such as shadows and reflections may not change consistently with the intended 3D edit (Fig.~\ref{fig:exp1}).

To reduce this ambiguity, another line of work lifts the edit into an explicit 3D space~\cite{liu2025shape}.
They reconstruct a 3D proxy, edit in this space, and then refine or re-render the result.
Such proxies can improve edit adherence when reconstruction is accurate, but they typically require costly per-video reconstruction and optimization, and are brittle under complex real-world dynamics (Fig.~\ref{fig:intro_fail}).
When the proxy is inaccurate or temporally unstable, its errors directly propagate to the edited output, making the result unreliable and secondary-effect updates inconsistent~\cite{liu2025shape}.

Together, these limitations point to a central challenge: object-level video geometric revision requires specifying the target object's 3D state change, while preserving non-edited regions, maintaining temporal coherence, and updating geometry-dependent secondary effects consistently.
Rather than using a fragile per-video reconstruction as an intermediate editing proxy, a scalable approach should learn a generative editor that directly executes the edit from lightweight geometric guidance.
This introduces a supervision challenge: learning such edits requires controlled before/after examples in which the intended geometric intervention is isolated, so the model can learn both the target transformation and its visual consequences.
Such controlled pairs are rare in real-world videos: unconstrained videos entangle object motion, camera motion, lighting variation, and scene dynamics, making it difficult to isolate the effects of the edit itself.
This lack of supervision limits scalable learning of faithful geometric editing and transformation-consistent secondary effects.

To overcome these challenges, we introduce \method, a unified geometry-instructed video editing framework that represents object-level geometric edits as transitions between pre-edit and post-edit object states.
Rather than relying on accurate per-video 3D reconstruction, which is costly and prone to error propagation, \method\ specifies these states with two lightweight, video-aligned geometry streams: a \textbf{depth-box} stream capturing coarse 3D placement and extent, and an \textbf{orientation-box} stream providing an appearance-agnostic heading cue.
Together, the pre/post streams define the source object to be transformed and the desired target state.
% , while avoiding leakage of fine-grained geometry or texture.
This object state transition formulation provides a shared operator-agnostic representation: the intended operation is induced by the difference between pre-edit and post-edit states, rather than specified through a task-specific label or control format.
It supports diverse DCC-style edits, including translation, rotation, scaling, duplication, removal, and trajectory editing, without requiring a fully reconstructed 3D proxy.
At inference time, these geometry streams are obtained from lightweight user input combined with off-the-shelf video segmentation and monocular geometry estimation (see Fig.~\ref{fig:teaser}), without requiring privileged engine buffers or per-video optimization.

The pre/post state formulation requires paired supervision where the state transition is isolated and secondary effects change consistently.
Such supervision can be obtained from a graphics engine that executes object-level edit programs and renders controlled before/after video pairs.
Specifically, our Unreal Engine pipeline samples combinatorial variations over scenes, assets, attributes (\textit{e.g.}, motion clips), camera trajectories, operator types, and operator parameters.
For each configuration, the engine executes the edit program while holding non-edit factors fixed.
Under physically based rendering, geometry-dependent effects such as shadows and reflections change \emph{consistently} with the transformation, providing direct supervision for faithful operator execution and non-edited content preservation.
This pipeline yields \dataset, a large-scale paired dataset for training, and \benchmark, a held-out reference-based benchmark for reproducible evaluation with well-defined targets.

Finally, \method\ demonstrates reliable geometric editing across controlled and in-the-wild evaluations.
On \benchmark, it outperforms representative baselines across all operators in PSNR/SSIM/LPIPS.
On real videos, it achieves the highest WinRate in our blinded user study, indicating stronger perceived edit faithfulness.
Qualitative comparisons (Figs.~\ref{fig:exp1} and~\ref{fig:exp2}) further show that \method\ better preserves non-edited regions and produces geometry-consistent secondary effects aligned with the intended transformations.
Additional analyses of robustness to estimated geometry, conditioning efficiency, realistic paired evaluation, more qualitative comparisons, and public-backbone instantiation are provided in the supplementary material.

% \vspace{-5pt}
\section{Related Work}
\label{sec:related}

\paragraph{Digital content creation via generative models.}
Recent image~\cite{ho2020denoising,song2021denoising,rombach2022high} and video generative models~\cite{ho2022imagen,blattmann2023stable,videoworldsimulators2024} are increasingly used in DCC workflows~\cite{luma2025dm,wan2025video,runway2025gen4,comanici2025gemini}, complementing traditional tools such as Blender~\cite{blender2025about} and Unreal Engine~\cite{unrealengine5} for iterative content revision.
This motivates diverse editing methods~\cite{zhang2023adding,ku2024anyv2v,ouyang2024i2vedit}, but video revision remains challenging because object-level edits must stay temporally coherent with scene structure.
\paragraph{Interactive and instruction-conditioned video editing and generation.}
Recent video editing and generation methods have explored increasingly flexible user controls, ranging from text prompts and region masks to point/trajectory guidance and reference-based signals.
Many works design task-specific interfaces for drag- or point-based manipulation~\cite{deng2024dragvideo,wu2024draganything,ma2023magicstick}, mask-guided inpainting and removal~\cite{zhou2023propainter,miao2025rose,bian2025videopainter,yang2025mtv}, layer decomposition~\cite{lee2025generative}, or trajectory- and motion-conditioned generation~\cite{gu2025diffusion,ma2024trailblazer,wang2025levitor,li2025magicmotion,chen2025perception,zhang2025motionpro,wang2025ati}.
Beyond operator-specific methods, recent general-purpose editors aim to unify multiple video creation and editing tasks within a single framework~\cite{ye2025unic,ju2025editverse,team2025kling,jiang2025vace}.
In contrast, \method\ targets source-video editing through 3D object-state transitions, providing a shared control format for changes in position, orientation, scale, presence, and trajectory while preserving non-edited content.

\paragraph{3D-aware image and video editing.}
3D cues provide an intuitive way to specify object-level intent under changing viewpoints.
In image editing, prior methods use synthetic 3D rendering, depth proxies, or coarse guidance for object transforms and manipulation~\cite{michel2023object,pandey2024diffusion,cheng20253d}.
Extending such control to videos further requires temporal coherence: VideoHandles~\cite{koo2025videohandles} lifts static scenes for training-free video composition, while Shape-for-Motion~\cite{liu2025shape} relies on time-consistent 3D proxies with per-video reconstruction/optimization.
In contrast, \method\ uses lightweight pre/post geometry instruction streams to enable 3D-aware source-video editing across diverse DCC-style operators without full per-video reconstruction.

\begin{figure}[t]
  \centering
  \includegraphics[width=\columnwidth]{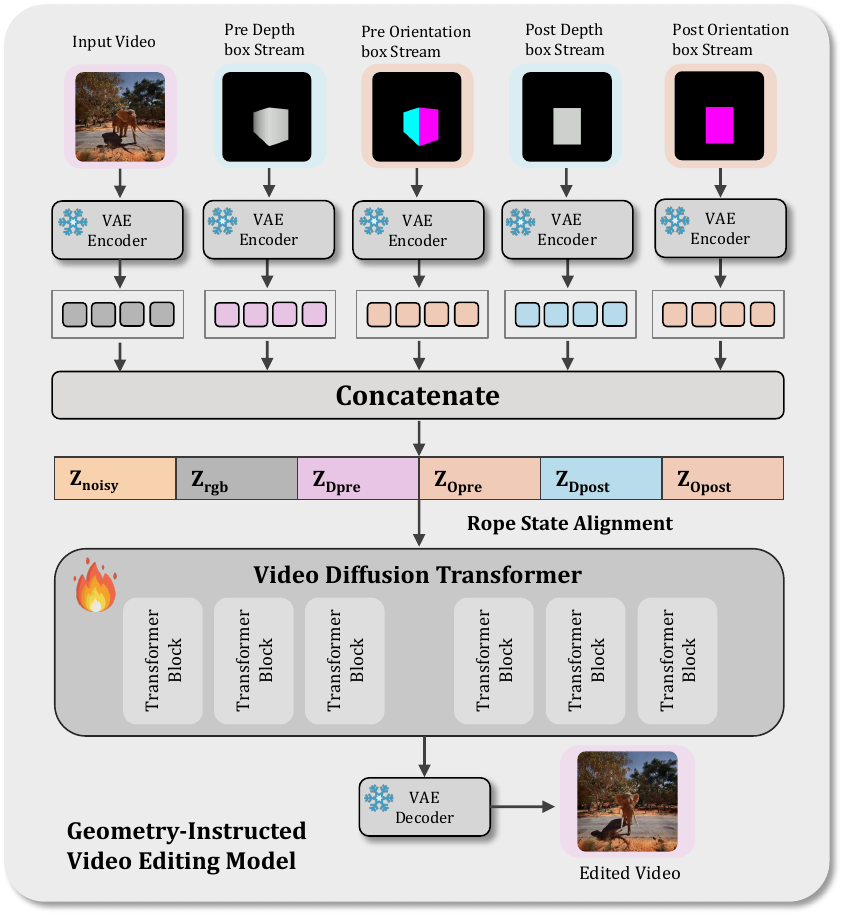}
  \vspace{-20pt}
  \caption{\textbf{Architecture of \method.}
  The input RGB video and four geometry-instruction streams are encoded by a shared video VAE into spatiotemporal tokens. These tokens are concatenated with the noisy edited-video latent and denoised by a video diffusion transformer with RoPE state alignment. The predicted edited latent is decoded to produce the edited video.}
  \label{fig:arch}
  \vspace{-10pt}
\end{figure}

\section{Overview}
\label{sec:method_overview}

We propose a \textbf{G}eometry-\textbf{I}nstructed \textbf{V}ideo \textbf{E}diting (\method) framework that exposes DCC-style object-level operators through transitions between pre/post object states. 
Given an input video $\mathbf{V}=\{\mathbf{I}_t\}_{t=1}^{T}$ and a geometric instruction $\mathbf{G}$, the editor produces: 
\begin{equation}
\hat{\mathbf{V}} = f_{\theta}\big(\mathbf{V},\, \mathbf{G}\big),
\end{equation}
where $\mathbf{G}$ denotes temporally aligned pre/post object-state streams, instantiated as depth-box and orientation-box streams (Sec.~\ref{sec:gi_streams}).
At inference time, \method\ requires only the input video and the finalized geometry instruction $(\mathbf{V}, \mathbf{G})$.
% the graphics engine is used only during training to generate paired supervision.
Learning this state representation further requires paired supervision in which the intended geometric intervention is isolated and its secondary effects are rendered consistently.
We therefore use a graphics engine to execute edit programs and generate controlled before/after pairs for training.
We instantiate $f_{\theta}$ as a conditional video diffusion model. The source video and the pre/post geometry streams are encoded into video-VAE latent tokens, and a diffusion denoiser synthesizes the edited-video latent conditioned on the source-video tokens and geometry-instruction (GI) tokens (Sec.~\ref{sec:model}). Our approach comprises two key components:

\textbf{(i) Unified object-state editing model (Sec.~\ref{sec:model}).}
\method\ provides a unified interface for DCC-style object-level editing by representing edits as transitions between pre-edit and post-edit object states. Each state is encoded using video-aligned geometry-instruction streams, which are later tokenized as geometry-instruction (GI) tokens to condition a video diffusion editor (Secs.~\ref{sec:gi_streams} and~\ref{sec:conditioning}). This pre/post formulation allows diverse edits to share the same control format; in this work, we instantiate it for translation, rotation, scaling, duplication, removal, and trajectory editing.

\textbf{(ii) Scalable procedural data synthesis (Sec. \ref{sec:data}).}
We use a graphics engine to execute object-level edit programs (Sec.~\ref{sec:edit_programs}) and render aligned before/after video pairs $(\mathbf{V}, \mathbf{V}')$ under diverse conditions. Physically based rendering keeps the edit as the intended intervention and allows geometry-dependent secondary effects to change consistently with the applied transformation.

\vspace{-5pt}
\section{Unified Geometry-instructed Video Editing Model}
\label{sec:model}
We now describe the unified geometry-instructed editing model, \method. The model is built around a compact pre/post object-state representation and a source-conditioned latent video diffusion editor. We first define the geometry-instruction streams used to encode pre-edit and post-edit object states, then describe how these streams are tokenized and integrated into the diffusion editor, and finally explain how the representation is instantiated for real-world videos.

\vspace{-8pt}
\subsection{Geometry Instruction Streams}
\label{sec:gi_streams}
Given the target object’s pre-edit and post-edit 3D oriented bounding boxes, we rasterize them into temporally aligned geometry instruction streams. These streams are later encoded into geometry-instruction (GI) tokens and used to condition the video diffusion model (Sec.~\ref{sec:conditioning}). We use two complementary instruction stream types, both derived from the underlying 3D box representation. 
The pre-edit streams anchor the edited object in the input video, while the post-edit streams specify the desired target state. The edit operator is therefore represented implicitly by the transition between these two states, rather than by a separate task-specific label.

\paragraph{Depth-box stream.}

The depth-box stream is a time-varying geometry instruction sequence of maps aligned to the video timeline,  encoding the target object’s coarse 3D placement and spatial extent in camera coordinates.
We rasterize the projected surfaces of the 3D oriented bounding box onto a black canvas using a z-buffer.
For each covered pixel, we write the nearest-surface depth and normalize it to $[0,1]$ using a scene-specific depth interval shared by the paired pre/post renders.
This ensures consistent normalization between pre/post streams within each training pair, while allowing different scenes to use different depth ranges.
Values outside the interval are clamped, and uncovered pixels remain zero.

\paragraph{Orientation-box stream.}
The orientation-box stream is a time-varying map sequence that encodes object heading in the current camera view in an appearance-agnostic manner.
We render the projected 3D box using fixed face-identity colors: each of the six faces is assigned a canonical RGB color shared across all samples. The resulting color pattern provides an anisotropic orientation cue while avoiding leakage of object texture or appearance. Since the face colors are fixed and globally shared, the orientation stream encodes pose only and cannot represent object identity.

\paragraph{Object-state formulation.}
Each training example therefore consists of five synchronized video streams: the input RGB video $\mathbf{V}$ and four geometry instruction streams describing the target object state before and after editing,
\begin{equation}
\mathcal{U} = \{\mathbf{V},\ \mathbf{V}^{d}_{\mathrm{pre}},\ \mathbf{V}^{o}_{\mathrm{pre}},\ \mathbf{V}^{d}_{\mathrm{post}},\ \mathbf{V}^{o}_{\mathrm{post}}\}.
\end{equation}
Here $(\cdot)^{d}$ and $(\cdot)^{o}$ denote the depth-box and orientation-box streams, respectively. All streams are rendered on the same timeline and are temporally aligned with $\mathbf{V}$. 
The edit operator is induced by the difference between the pre-edit and post-edit states, rather than provided as a separate operator-specific input.
Special cases are handled naturally as object-state transitions. Object removal is represented by empty post-edit streams rendered as all-black canvases, indicating a transition from an existing pre-edit object state to an absent post-edit state. For duplication, multiple spawned instances specified by the edit program are rasterized into the post-edit streams, representing a transition from one source object state to multiple target states, with occlusions handled automatically by the z-buffer.

\subsection{Editor Video Diffusion Model}
\label{sec:conditioning}
We instantiate the editor as a conditional latent video diffusion model. Given the source video $\mathbf{V}$, the noisy edited latent $\tilde{\mathbf{z}}'_{\tau}$ at timestep $\tau$, and the pre/post geometry instruction streams, we encode all streams using a shared video VAE and flatten the resulting spatiotemporal latents into token sequences. 
Let $\mathbf{S}_{\mathrm{tgt}}$ denote the token sequence obtained from the noisy edited latent $\tilde{\mathbf{z}}'_{\tau}$, and $\mathbf{S}_{\mathrm{src}}$ the token sequence from the source video $\mathbf{V}$. We concatenate these with the pre-edit and post-edit GI token blocks as: 
\begin{equation}
\tilde{\mathbf{S}} = [\,\mathbf{S}_{\mathrm{tgt}};\ \mathbf{S}_{\mathrm{src}};\ \mathbf{S}_{\mathrm{pre}};\ \mathbf{S}_{\mathrm{post}}\,].
\label{eq:seq_concat}
\end{equation}

We refer to tokens originating from the auxiliary streams as geometry-instruction (GI) tokens.
Sequence-level concatenation preserves stream identity and allows self-attention to compare pre/post state tokens at corresponding spatiotemporal locations, enabling the model to infer the intended edit from relations between object states rather than an explicit operator label.
For efficiency, we temporally downsample the GI streams before tokenization using stride $s=2$ by default, and align their RoPE indices to the corresponding source-video timestamps.
The concatenated sequence is then processed by a video diffusion transformer as shown in Fig.~\ref{fig:arch}.

\paragraph{RoPE state alignment.}
A naive concatenation in Eq.~(\ref{eq:seq_concat}) would assign independent positional indices to different blocks, causing tokens from different streams to appear unrelated under rotary positional encoding (RoPE), even when they correspond to the same spatiotemporal location. To address this, we align RoPE indices across all streams by tying them to the latent-grid coordinate.

After VAE encoding, all streams share the latent-grid resolution $(T_\ell,H_\ell,W_\ell)$.
Let a token correspond to coordinate $(t,h,w)$, where $t\in\{0,\dots,T_\ell-1\}$, $h\in\{0,\dots,H_\ell-1\}$, and $w\in\{0,\dots,W_\ell-1\}$.
We define a canonical linearized index:

\begin{equation}
\phi(t,h,w) = t\cdot H_\ell W_\ell + h\cdot W_\ell + w .
\end{equation}
RoPE is applied using $\phi(t,h,w)$ for tokens from all streams, ensuring that tokens corresponding to the same spatiotemporal location share the same base positional encoding.

To distinguish source-state tokens from target-state tokens, we introduce a constant state offset. Tokens corresponding to the edited target, namely the noisy edited latent and the post-edit GI tokens, are assigned offset $0$. Tokens corresponding to the source video and pre-edit GI streams receive an additional offset $\Delta_{\mathrm{state}}$: 
\begin{equation}
p = \phi(t,h,w) + \Delta_{\mathrm{state}}\cdot \mathbb{I}[\text{source-state}].
\end{equation}
This preserves spatiotemporal correspondence across streams while encoding the semantic distinction between source and target states.

\paragraph{Training objective.}
We follow the rectified-flow objective of the pretrained video backbone.
Given a rendered pair $(\mathbf{V},\mathbf{V}')$ and instruction $\mathbf{G}$, the diffusion transformer predicts the velocity of the noised edited-video latent, conditioned on the source-video latent and the instruction.
The source-video latent provides appearance and background context, while $\mathbf{G}$ specifies the intended pre/post state transition.
The full objective is provided in the supplement.

\begin{figure*}[t]
\centering
\includegraphics[width=0.95\textwidth]{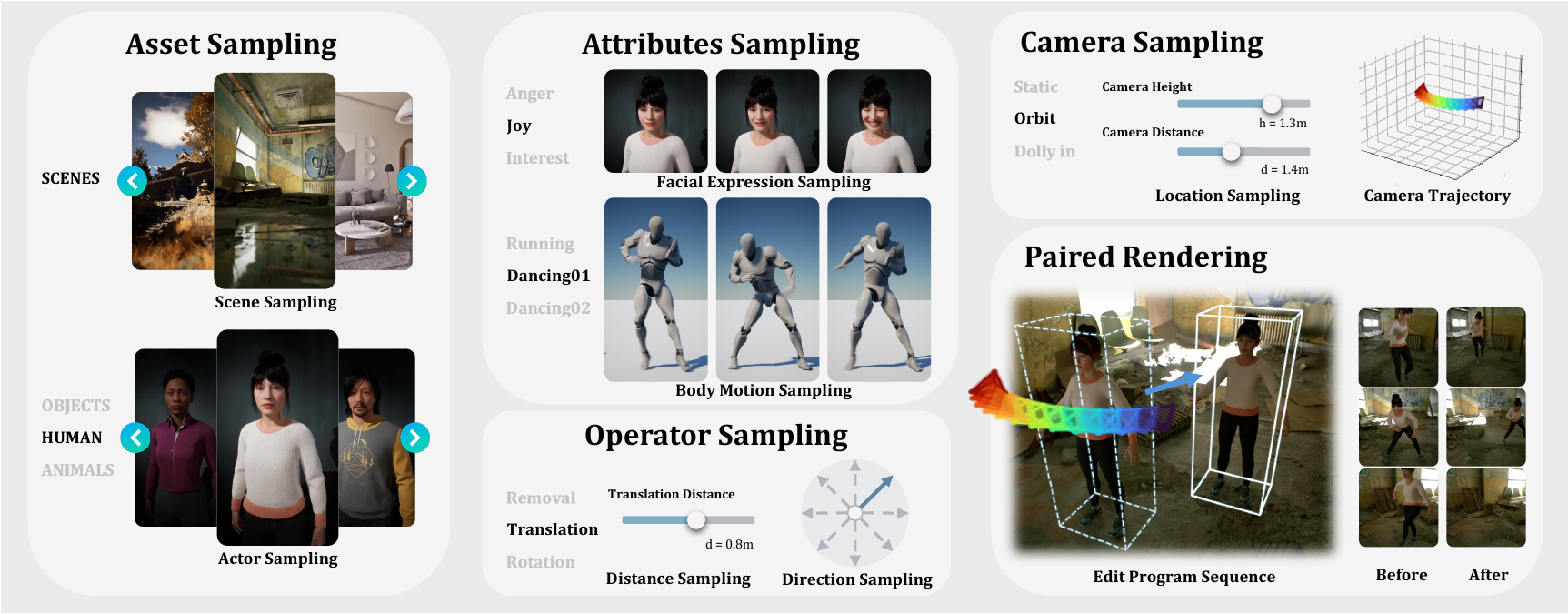}
% \vspace*{-3.5mm}
\vspace{-10pt}
\caption{\textbf{Procedural Construction Pipeline.}
From left to right, (i) \textit{Asset Sampling}: selection of a scene and a target actor;
(ii) \textit{Attributes Sampling}: assignment of actor attributes independent of the edit (e.g., facial expressions and body-motion clips);
(iii) \textit{Camera Sampling}: sampling of a camera-motion family and continuous parameters to define the camera trajectory;
(iv) \textit{Operator Sampling}: sampling of a DCC-style operator and its parameters (e.g., translation direction and distance) to instantiate an edit program; and
(v) \textit{Paired Rendering}: execution of the edit program and rendering under identical camera and environment settings, producing a controlled before/after video pair where the intended manipulation is the primary change.}
\label{fig:engine_recipe}
\vspace{-10pt}
\end{figure*}

\subsection{Geometry-instructed Editing for Real Videos}
\label{sec:instruction_acquisition}
At inference time, we extend our geometry-instruction interface to real videos by converting a user-selected target in $\mathbf{V}$ and an edit operator into temporally consistent GI streams. We use off-the-shelf tools~\cite{ravi2024sam,jiang2025geo4d} to recover a 3D bounding box trajectory and an orientation reference for the target object.

Specifically, we first obtain a spatiotemporally consistent object mask using SAM 2. The segmented pixels are then back-projected into 3D using estimated depth maps and camera poses, from which we fit a temporally aligned 3D bounding box over the video. 
For orientation, we do not estimate an independent object orientation for every frame, as such estimates are often unstable in casual videos.
Instead, the user specifies a reference heading in one frame, which serves as a canonical heading anchor.
The pre-edit orientation stream is generated by propagating this anchor through the recovered camera poses and 3D box trajectory, and rasterizing the oriented boxes under the observed viewpoints.
The post-edit orientation stream is generated in the same way after applying the user-specified relative 3D rotation to the anchor.
Thus, the pre/post streams encode the intended relative heading change with respect to a shared anchor, mainly reflecting viewpoint changes and applied edits rather than noisy motion-induced orientation variations.

When automatic geometry estimates are noisy, we allow lightweight correction of the fitted box trajectory before generation. The finalized geometry instruction is then fixed and rasterized into the same instruction-stream format used during training, enabling geometry-instruction (GI) token conditioning at inference time without requiring engine buffers or per-video optimization.

\section{Procedural Data Synthesis}
\label{sec:data}
A key challenge in learning object-level video editing is the lack of paired supervision in which the intended geometric edit is clearly defined and secondary effects evolve consistently with that edit.
We address this with a scalable procedural synthesis pipeline that executes explicit object-level edit programs in a graphics engine and renders controlled before/after video pairs.

\vspace{-5pt}
\subsection{Engine-Rendered Paired Data}
\label{sec:edit_programs}
We render paired videos $(\mathbf{V}, \mathbf{V}')$ under matched scene, camera, and rendering settings, so that the edit program is the only intended intervention.
The graphics engine is used only to generate paired supervision during training; at inference time, the model operates solely on the input video $\mathbf{V}$ and the geometry instruction $\mathbf{G}$.

We represent user intent as an \emph{edit program} acting on a target object $o_k$, which produces a time-varying edited object state. At each timestep $t$, the object state is parameterized as
\begin{equation}
\mathcal{X}_{k,t} = \big(\mathbf{T}_{k,t},\, \boldsymbol{\sigma}_{k,t},\, \gamma_{k,t}\big),
\end{equation}
where $\mathbf{T}_{k,t} \in SE(3)$ denotes the rigid pose mapping object-local coordinates to world coordinates, $\boldsymbol{\sigma}_{k,t} \in \mathbb{R}_+^3$ represents anisotropic scale, and $\gamma_{k,t} \in \{0,1\}$ indicates object presence.

An edit program specifies time-varying transformations: 
\begin{equation}
\mathbf{P} = \Big(\Delta \mathbf{T}(t),\, \Delta \boldsymbol{\sigma}(t),\, m(t),\, \mathcal{D}\Big), \quad t \in \{1,\dots,T\},
\end{equation}
which are applied to the object state as: 
\begin{equation}
\begin{aligned}
\mathbf{T}'_{k,t} = \mathbf{T}_{k,t}\,\Delta\mathbf{T}(t),
\boldsymbol{\sigma}'_{k,t} = \Delta\boldsymbol{\sigma}(t)\odot \boldsymbol{\sigma}_{k,t},
\gamma'_{k,t} = m(t)\,\gamma_{k,t}.
\end{aligned}
\label{eq:apply}
\end{equation}
$\Delta\mathbf{T}(t)$ is defined in the object’s local coordinate frame, following standard DCC conventions. This formulation naturally supports translation and rotation via $\Delta\mathbf{T}(t)$, scaling via $\Delta\boldsymbol{\sigma}(t)$, removal via the masking function $m(t)$, duplication via a spawn function $\mathcal{D}$, and trajectory editing through time-varying transformations.

Given an instantiated edit program, the engine renders a paired sample consisting of the source video $\mathbf{V}$, the edited video $\mathbf{V}'$, and the corresponding pre-edit and post-edit geometry instruction streams.  Each rendered example yields a synchronized tuple: 
\begin{equation}
\big(\mathbf{V},\mathbf{V}', \mathbf{V}^{d}_{\mathrm{pre}}, \mathbf{V}^{o}_{\mathrm{pre}}, \mathbf{V}^{d}_{\mathrm{post}}, \mathbf{V}^{o}_{\mathrm{post}}\big), 
\end{equation}
which serves as supervision for diffusion training and evaluation.

\subsection{Procedural Construction Pipeline}
\label{sec:engine_teacher_pipeline}

We now describe how the engine-rendered paired supervision is instantiated at scale through a procedural construction pipeline. It operationalizes the edit-program by systematically sampling scene content, object attributes, camera trajectories, and edit parameters, and executing the resulting edit programs in the graphics engine.

The pipeline is inherently scalable. Data generation is decomposed into independent, programmatic components that can be sampled and combined without manual intervention. Each pair is produced by executing a lightweight edit program on a fixed scene setup, yielding linear scaling with the number of renders and eliminating the need for human annotation. Moreover, the process is parallel across scenes, operators, and parameter configurations, enabling large-scale synthesis using standard render farms.

Fig.~\ref{fig:engine_recipe} provides an overview of the procedural construction pipeline.
Each training example is generated by composing a fixed scene setup with a sampled edit program.
We first perform \emph{Asset Sampling} to select a scene and target object, \emph{Attribute Sampling} to assign edit-independent factors such as motion clips, and \emph{Camera Sampling} to choose a camera-motion family (e.g., orbit, dolly) and its parameters.

Given this fixed setup, we sample a DCC-style operator and its parameters to instantiate an edit program $\mathbf{P}$ as defined in Eq.~\eqref{eq:apply}. The engine executes $\mathbf{P}$ on the target object state over time and performs \emph{Paired Rendering} to produce the source and edited videos together with corresponding pre-edit and post-edit geometry instruction streams, $(\mathbf{V}, \mathbf{V}^{d}_{\mathrm{pre}}, \mathbf{V}^{o}_{\mathrm{pre}})$ and $(\mathbf{V}', \mathbf{V}^{d}_{\mathrm{post}}, \mathbf{V}^{o}_{\mathrm{post}})$. 
All streams are rendered at the same resolution and temporally aligned, while keeping the scene, camera trajectory, renderer settings, and post-processing fixed so that $\mathbf{P}$ is the only intended intervention.
Under physically based rendering, geometry-dependent effects update consistently with the executed transformation, yielding a clean supervision for faithful edits and preservation of non-edited content.

Finally, the pipeline is operator-agnostic within our object-state formulation. New DCC-style editing capabilities that can be expressed as object-state changes can be incorporated by defining additional edit programs, without modifying the rendering interface or geometry-instruction stream representation. This modular design enables continuous expansion of the dataset while preserving consistent supervision and unified conditioning.

\begin{table}[t]
\caption{\textbf{Quantitative comparison on \benchmark.}
We report PSNR/SSIM/LPIPS against the rendered reference $\mathbf{V}'$ for each operator.}
\label{tab:main_quant_psnr}
\vspace{-4pt}
\centering
\scriptsize
\setlength{\tabcolsep}{3pt}
\renewcommand{\arraystretch}{1.0}

\begin{tabularx}{0.9\columnwidth}{l l YYY}
\toprule
Task & Method
& \multicolumn{1}{c}{\thead{PSNR$\uparrow$}}
& \multicolumn{1}{c}{\thead{SSIM$\uparrow$}}
& \multicolumn{1}{c}{\thead{LPIPS$\downarrow$}} \\
\midrule

\multirow{3}{*}{Rotation}
& DragVideo                        & 16.72 & 0.5601 & 0.4109 \\
& Kling-O1                         & 19.49 & 0.7151 & 0.1793 \\
& \cellcolor{tabhighlight}\textbf{\method\ (Ours)}
& \cellcolor{tabhighlight}\textbf{21.66}
& \cellcolor{tabhighlight}\textbf{0.7848}
& \cellcolor{tabhighlight}\textbf{0.1558} \\
\midrule

\multirow{4}{*}{Removal}
& ProPainter                       & 25.79 & 0.7958 & 0.1252 \\
& VACE                             & 23.38 & 0.7439 & 0.1337 \\
& ROSE                             & 27.15 & 0.8103 & 0.1283 \\
& \cellcolor{tabhighlight}\textbf{\method\ (Ours)}
& \cellcolor{tabhighlight}\textbf{27.33}
& \cellcolor{tabhighlight}\textbf{0.8433}
& \cellcolor{tabhighlight}\textbf{0.1201} \\
\midrule

\multirow{3}{*}{Scaling}
& DragVideo                        & 15.81 & 0.3467 & 0.3193 \\
& Kling-O1                         & 16.70 & 0.5485 & 0.2329 \\
& \cellcolor{tabhighlight}\textbf{\method\ (Ours)}
& \cellcolor{tabhighlight}\textbf{20.11}
& \cellcolor{tabhighlight}\textbf{0.7498}
& \cellcolor{tabhighlight}\textbf{0.1648} \\
\midrule

\multirow{3}{*}{Translation}
& DragVideo                        & 14.76 & 0.3337 & 0.4438 \\
& Kling-O1                         & 16.66 & 0.5572 & 0.2769 \\
& \cellcolor{tabhighlight}\textbf{\method\ (Ours)}
& \cellcolor{tabhighlight}\textbf{19.97}
& \cellcolor{tabhighlight}\textbf{0.7348}
& \cellcolor{tabhighlight}\textbf{0.2272} \\
\midrule

\multirow{2}{*}{Duplication}
& Kling-O1                         & 14.59 & 0.5620 & 0.3262 \\
& \cellcolor{tabhighlight}\textbf{\method\ (Ours)}
& \cellcolor{tabhighlight}\textbf{19.72}
& \cellcolor{tabhighlight}\textbf{0.7891}
& \cellcolor{tabhighlight}\textbf{0.1275} \\
\midrule

\multirow{2}{*}{Trajectory Editing}
& DaS                              & 22.60 & 0.7458 & 0.1788 \\
& \cellcolor{tabhighlight}\textbf{\method\ (Ours)}
& \cellcolor{tabhighlight}\textbf{27.41}
& \cellcolor{tabhighlight}\textbf{0.9019}
& \cellcolor{tabhighlight}\textbf{0.0681} \\
\bottomrule
\end{tabularx}
\vspace{-15pt}
\end{table}

\vspace{-5pt}
\subsection{\dataset}
\label{sec:dataset_instantiation}
We instantiate the above pipeline to construct \dataset, a large-scale paired video dataset for object-level geometric editing. Each example in \dataset\ contains a paired supervision target $(\mathbf{V}, \mathbf{V}')$ together with aligned geometry instruction streams (Sec.~\ref{sec:gi_streams}); more details are provided in the supplementary material.

To promote robustness across content, motion, and viewpoint, we sample a diverse set of scenes and assets, including indoor and outdoor environments and object categories such as humans, animals, vehicles, and rigid props. For articulated subjects, we attach motion clips during rendering, while rigid objects remain static; relative motion arises from camera movement and, when applicable, the geometric edit itself. We include multiple camera-motion families to vary viewpoint while preserving paired control.

We instantiate \dataset\ using a representative set of DCC-style operators with explicit geometric targets, including rotation, translation, scaling, removal, duplication, and trajectory editing. For each example, we sample operator parameters 
% (e.g., angles, offsets, scale factors, duplication counts) 
to form diverse edit programs. Sampling is balanced across asset categories, while operator frequencies are chosen to ensure sufficient coverage and stable training. Lightweight validity checks filter rare degenerate cases, such as severe target invisibility or invalid projections.

In total, \dataset\ contains 202k paired clips across 30 scenes and 319 target instances, rendered at 100 frames per clip and $1024\times1024$ resolution. It includes 56k rotation, 49k translation, 32k scaling, 45k removal, 5k duplication, and 15k trajectory-editing pairs.
\dataset\ is designed primarily to provide controlled paired supervision for object-state transitions, rather than to replace the broad visual prior of the pretrained video backbone.
Its diversity arises from combinatorial sampling of scenes, target categories, articulated motions, camera trajectories, operator types, and operator parameters.
These paired examples teach the editor how pre/post geometry instructions map to executable object-level edits, while broad appearance and motion priors are inherited from the pretrained backbone.

\begin{table}[t]
\caption{\textbf{Real-video evaluation.}
We report aggregated VBench scores and best-of-$K_\ell$ user-study WinRate. Temporal averages Temporal Flickering and Motion Smoothness; Consistency averages Subject and Background Consistency; Quality averages Imaging and Aesthetic Quality.}
\label{tab:main_quant_inwild_compact_1us}
\vspace{-4pt}
\centering
\scriptsize
\setlength{\tabcolsep}{2.5pt}
\renewcommand{\arraystretch}{1}

\begin{tabularx}{0.98\columnwidth}{l l YYY Y}
\toprule
\multirow{2}{*}{Task} & \multirow{2}{*}{Method}
& \multicolumn{3}{c}{\textbf{Video Quality}}
& \multicolumn{1}{c}{\textbf{User Study}} \\
\cmidrule(lr){3-5}\cmidrule(lr){6-6}
& &
\multicolumn{1}{c}{\thead{Temporal$\uparrow$}} &
\multicolumn{1}{c}{\thead{Consistency$\uparrow$}} &
\multicolumn{1}{c}{\thead{Quality$\uparrow$}} &
\multicolumn{1}{c}{\thead{WinRate$\uparrow$}} \\
\midrule

\multirow{3}{*}{Rot.}
& DragVideo
& 0.9873 & 0.9451 & 0.5728 & 0.58\% \\
& Kling-O1
& 0.9915 & 0.9761 & \textbf{0.6574} & 32.67\% \\
& \cellcolor{tabhighlight}\textbf{\method\ (Ours)}
& \cellcolor{tabhighlight}\textbf{0.9923}
& \cellcolor{tabhighlight}\textbf{0.9775}
& \cellcolor{tabhighlight}0.6257
& \cellcolor{tabhighlight}\textbf{66.75\%} \\
\addlinespace[1pt]
\midrule

\multirow{4}{*}{Rem.}
& ProPainter
& 0.9862 & 0.9587 & 0.5188 & 1.41\% \\
& VACE
& 0.9795 & 0.9488 & 0.5079 & 0.42\% \\
& ROSE
& 0.9894 & 0.9679 & 0.5106 & 21.75\% \\
& \cellcolor{tabhighlight}\textbf{\method\ (Ours)}
& \cellcolor{tabhighlight}\textbf{0.9910}
& \cellcolor{tabhighlight}\textbf{0.9742}
& \cellcolor{tabhighlight}\textbf{0.5302}
& \cellcolor{tabhighlight}\textbf{76.42\%} \\
\addlinespace[1pt]
\midrule

\multirow{3}{*}{Scale}
& DragVideo
& 0.9893 & 0.9546 & 0.6123 & 0.67\% \\
& Kling-O1
& 0.9938 & \textbf{0.9792} & \textbf{0.6631} & 16.75\% \\
& \cellcolor{tabhighlight}\textbf{\method\ (Ours)}
& \cellcolor{tabhighlight}\textbf{0.9944}
& \cellcolor{tabhighlight}0.9738
& \cellcolor{tabhighlight}0.6560
& \cellcolor{tabhighlight}\textbf{82.58\%} \\
\addlinespace[1pt]
\midrule

\multirow{3}{*}{Trans.}
& DragVideo
& 0.9829 & 0.9587 & 0.5977 & 1.50\% \\
& Kling-O1
& \textbf{0.9907} & 0.9722 & 0.6214 & 22.83\% \\
& \cellcolor{tabhighlight}\textbf{\method\ (Ours)}
& \cellcolor{tabhighlight}0.9893
& \cellcolor{tabhighlight}\textbf{0.9753}
& \cellcolor{tabhighlight}\textbf{0.6296}
& \cellcolor{tabhighlight}\textbf{75.61\%} \\
\addlinespace[1pt]
\midrule

\multirow{2}{*}{Dupl.}
& Kling-O1
& \textbf{0.9889} & \textbf{0.9724} & \textbf{0.6233} & 18.17\% \\
& \cellcolor{tabhighlight}\textbf{\method\ (Ours)}
& \cellcolor{tabhighlight}0.9872
& \cellcolor{tabhighlight}0.9683
& \cellcolor{tabhighlight}0.6017
& \cellcolor{tabhighlight}\textbf{81.83\%} \\
\addlinespace[1pt]
\midrule

\multirow{2}{*}{Traj.}
& DaS
& 0.9871 & 0.9431 & 0.5471 & 14.08\% \\
& \cellcolor{tabhighlight}\textbf{\method\ (Ours)}
& \cellcolor{tabhighlight}\textbf{0.9879}
& \cellcolor{tabhighlight}\textbf{0.9680}
& \cellcolor{tabhighlight}\textbf{0.5682}
& \cellcolor{tabhighlight}\textbf{85.92\%} \\
\bottomrule
\end{tabularx}

\vspace{-10pt}
\end{table}

\section{Experiments}
\label{sec:experiments}

\subsection{Datasets, Benchmark and Metrics}
\label{sec:exp_data_bench_metrics}

\paragraph{Datasets and Benchmarks.}
We train on \dataset\ generated by our procedural pipeline (Sec.~\ref{sec:data}), where each example provides a paired target $(\mathbf{V},\mathbf{V}')$ and aligned pre/post GI streams (Sec.~\ref{sec:gi_streams}).
For reference-based evaluation, we introduce \benchmark, a curated benchmark of paired edits rendered by UE5 under controlled settings.
\benchmark\ contains 30 clips for each operator, with scenes and target asset instances disjoint from \dataset.
To assess transfer to real videos, we evaluate on 240 in-the-wild clips from~\cite{Pexels} without paired references, with instructions obtained via user specification, SAM~2, and Geo4D (Sec.~\ref{sec:instruction_acquisition}).
Any manual correction is restricted to geometry estimation before method execution, and the finalized instruction is shared across methods.
We report VBench scores and a blinded best-of-$K_\ell$ user study with 30 participants; more details are provided in the supplementary material.

\paragraph{Evaluation metrics.}
We evaluate in two complementary settings:
(i) Reference-based metrics on \benchmark.
We report PSNR, SSIM, and LPIPS~\cite{ssim,zhang2018unreasonable} between the edited output $\hat{\mathbf{V}}$ and the rendered reference $\mathbf{V}'$ (Tab.~\ref{tab:main_quant_psnr}).
When methods output different lengths, we resample outputs to a common $T_\ell$ per operator $\ell$ before computing metrics.
(ii) Reference-free metrics and human evaluation on real videos. 
We report aggregated VBench scores~\cite{huang2023vbench} as reference-free measures of generic video quality (Tab.~\ref{tab:main_quant_inwild_compact_1us}). 
As these metrics do not directly measure edit faithfulness, we use them as complementary evidence and leverage a blinded best-of-$K_\ell$ user study to evaluate perceived edit adherence.

\begin{table}[t]
\caption{\textbf{Component ablation.}
We ablate the orientation-box tokens $z_o$, depth-box tokens $z_D$, and RoPE state alignment on \benchmark.}
\label{tab:ablation_components}
\vspace{-4pt}
\centering
\small
\setlength{\tabcolsep}{3pt}
\renewcommand{\arraystretch}{1.0}

\begin{tabularx}{0.8\columnwidth}{YYY YYY}
\toprule
\multicolumn{1}{c}{\thead{$z_o$}} &
\multicolumn{1}{c}{\thead{$z_D$}} &
\multicolumn{1}{c|}{\thead{RoPE state alignment}} &
\multicolumn{1}{c}{\thead{PSNR$\uparrow$}} &
\multicolumn{1}{c}{\thead{SSIM$\uparrow$}} &
\multicolumn{1}{c}{\thead{LPIPS$\downarrow$}} \\
\midrule
\multicolumn{1}{c}{\cmark} & \multicolumn{1}{c}{ }       & \multicolumn{1}{c|}{ }       & 20.93 & 0.7662 & 0.1952 \\
\multicolumn{1}{c}{ }       & \multicolumn{1}{c}{\cmark} & \multicolumn{1}{c|}{ }       & 19.67 & 0.7247 & 0.2313 \\
\multicolumn{1}{c}{\cmark} & \multicolumn{1}{c}{\cmark} & \multicolumn{1}{c|}{ }       & 21.89 & 0.7811 & 0.1779 \\
\multicolumn{1}{c}{\cmark} & \multicolumn{1}{c}{\cmark} & \multicolumn{1}{c|}{\cmark} & \textbf{22.70} & \textbf{0.8006} & \textbf{0.1439} \\
\bottomrule
\end{tabularx}

\vspace{-15pt}
\end{table}

\vspace{-5pt}
\subsection{Implementation Details}
\label{sec:exp_impl}

\paragraph{Training Details.}
\method\ is built by fine-tuning a pretrained transformer based latent video generation backbone with approximately 1B parameters, rather than training a model from scratch on \dataset.
We fine-tune the denoiser on \dataset\ for 24k steps using AdamW with a learning rate of $1\times10^{-5}$ and batch size 64. 
Additional backbone and optimization details are provided in the supplement.

\vspace{-5pt}
\paragraph{Evaluation Details.}
At inference, we condition on the input RGB video $\mathbf{V}$ and its geometry instruction streams.
On \benchmark, we use engine-rendered instruction streams (oracle GI streams).
For real videos, we obtain the target information via user interaction and estimate geometry using off-the-shelf tools as described in Sec.~\ref{sec:instruction_acquisition}.

\vspace{-5pt}
\paragraph{Baselines.}
Our benchmark targets \emph{video editing}: each method is evaluated on the same input video and the same intended edit, with the goal of preserving non-edited content.
We therefore prioritize baselines that natively operate on source videos, and use the closest executable interface when no fully matched method exists for an operator.
Accordingly, we compare with two categories of baselines: general-purpose video editors that support the target edit through their native interface, and operator-specific methods with the closest task overlap and executable interface.
For methods with different interfaces, we convert the same edit intent into the required input form.
All methods are run with official settings (see more details in the supplement). 
For general geometric edits, we include Kling-O1, a commercial general-purpose video editor, and DragVideo, a video-input drag-based editor, for rotation, scaling, and translation where drag-style controls can approximate the intended geometric edits.
For removal, we compare with representative video inpainting/removal methods ProPainter and ROSE, using the same target masks.
We also evaluate VACE on removal, where its unified editing interface most directly overlaps with our benchmark.
For trajectory editing, fully matched video-input editing baselines are limited, as most trajectory-conditioned methods are designed for generation rather than editing.
We therefore compare with DaS, a representative 3D-aware trajectory-conditioned video generation method, using point tracks derived from our geometry instructions.

\vspace{-5pt}
\subsection{Main Results and Ablation Studies}
\label{sec:exp_main}

\paragraph{Quantitative results.}
On \benchmark, Tab.~\ref{tab:main_quant_psnr} shows that \method\ consistently achieves higher PSNR/SSIM and lower LPIPS than baselines across all operators, reflecting more faithful edit under the controlled paired setting.
The largest gains appear on strongly geometry-dependent edits such as scaling and translation, where baselines tend to drift or under-edit.
On the real-video set, Tab.~\ref{tab:main_quant_inwild_compact_1us} reports reference-free VBench scores together with a blinded user study.
\method\ remains competitive on VBench scores, and often improves consistency metrics, indicating that stronger geometric control does not come at the cost of degraded video quality.
More importantly, \method\ achieves the highest WinRate across all operators, indicating stronger perceived edit faithfulness in the wild.

\begin{figure*}[p]
  \centering
  \includegraphics[width=\textwidth,height=0.95\textheight,keepaspectratio]{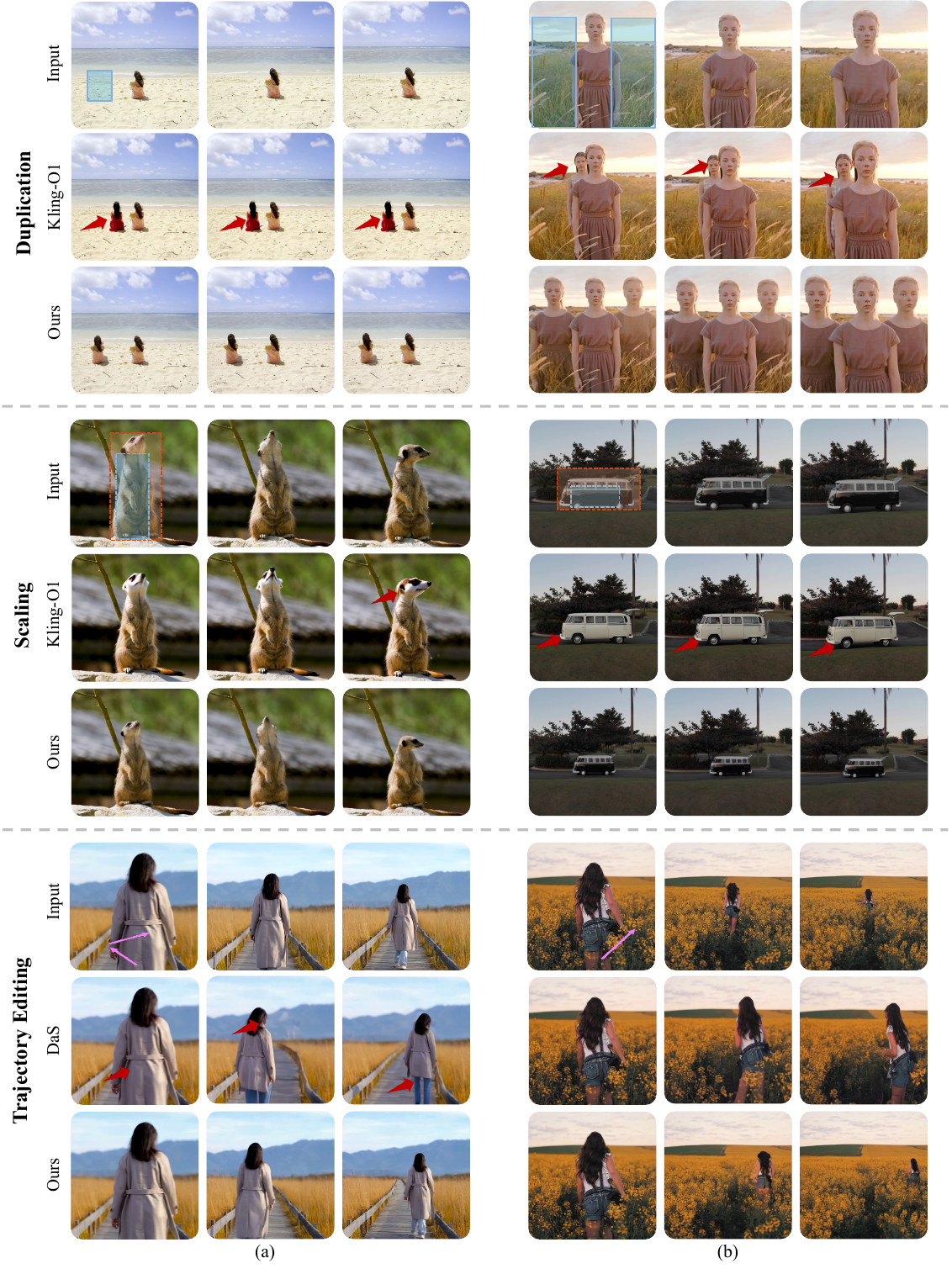}
  \vspace{-6pt}
  \caption{\textbf{Qualitative comparison on real videos (II).}
We show results for geometric operators including duplication, scaling and trajectory editing on the real videos.
\method\ consistently achieves more faithful geometric edits with stronger temporal coherence, while baselines may introduce identity/background drift, operator mis-execution, or temporally inconsistent artifacts. For trajectory editing, arrows in the input indicate the intended trajectory direction of the target.
}
  \label{fig:exp2}
\end{figure*}

\paragraph{Qualitative results.}
% Fig.~\ref{fig:exp1} and Fig.~\ref{fig:exp2} show representative real-video results.
Figs.~\ref{fig:exp1} and~\ref{fig:exp2} show representative real-video results, with one representative baseline per operator due to space constraints; extended operator-wise comparisons are provided in the supplement.
Across operators, \method\ better follows the intended geometric edit, preserves object identity/motion and non-edited regions, and maintains geometry-dependent effects such as shadows and reflections.
In contrast, baselines often mis-execute the operator (e.g., translation (b)), miss coherent secondary-effect updates (e.g., removal (a)), or introduce identity/background drift and temporal artifacts (e.g., scaling (a,b) and rotation (b)). 

\paragraph{Ablations.}
Tab.~\ref{tab:ablation_components} ablates the key components of \method.
Using both orientation-box and depth-box tokens outperforms either alone, indicating that the two streams provide complementary geometric cues.
Without orientation-box tokens, the model suffers from heading ambiguity, especially for rotation where the intended 3D turning direction becomes under-specified.
Without depth-box tokens, spatial grounding degrades, reducing accuracy for 3D position and scale changes.
RoPE state alignment further improves PSNR/SSIM/LPIPS by tying tokens at corresponding spatiotemporal locations across streams while preserving the distinction between pre/post states.

\section{Discussion and Conclusion}
\label{sec:conclusion}
We present \method, a geometry-instructed video editing framework for DCC-style object-level operations, built on a unified object-state representation and scalable procedural paired supervision.
By encoding geometric intent with lightweight depth-box and orientation-box streams, \method\ enables a single diffusion-based editor to support translation, rotation, scaling, duplication, removal, and trajectory editing, while maintaining temporal coherence and geometry-consistent secondary effects.
Evaluations on \benchmark\ and real-world videos demonstrate more faithful geometric editing than existing baselines, including commercial models.

A current limitation of \method\ is computational efficiency: 
% token-level conditioning concatenates multiple geometry-instruction streams with video tokens, increasing transformer sequence length and latency.
concatenating multiple geometry-instruction streams with video tokens increases sequence length and latency.
Reducing this overhead through more compact geometric representations or structured conditioning mechanisms is an important direction for future work.

% ----------------- One-page figure -----------------

% ---------------------------------------------------

% \clearpage
\appendix
\section*{\textbf{Supplementary Material}}

\renewcommand{\thefigure}{S\arabic{figure}}
\renewcommand{\thetable}{S\arabic{table}}
\setcounter{figure}{0}
\setcounter{table}{0}

\section{Base Text-to-Video Diffusion Model}

\subsection{Backbone details}
\method\ is built upon a transformer-based latent diffusion text-to-video (T2V) model. The model first compresses input videos from pixel space into a compact spatiotemporal latent representation using a 3D video VAE. Diffusion is then performed in this latent space with a transformer that operates on flattened spatiotemporal tokens.
A key design choice is to model space and time jointly in attention. Many prior video diffusion backbones adopt factorized architectures, e.g., coupling spatial attention with an extra 1D temporal attention module, or use designs inherited from image diffusion with ad-hoc temporal components. In our base model, we instead employ 3D self-attention over spatiotemporal tokens throughout the network, allowing each token to attend across both spatial locations and time steps within a unified mechanism. This yields stronger temporal coherence and higher visual fidelity in practice, and also provides a clean foundation for injecting geometry-conditioned signals in our editing setting.
For diffusion conditioning, we use standard timestep modulation in transformer blocks. Specifically, the diffusion timestep embedding is mapped to per-block modulation parameters that scale the normalized token features, and the resulting modulated tokens are fed into the attention and feed-forward sublayers. We adopt RMSNorm for token normalization, which we found stable for large-scale spatiotemporal transformers.

\subsection{Training objective}
Given a rendered training pair $(\mathbf{V}, \mathbf{V}')$ and its instruction representation $\mathbf{G}$, we train the conditional latent video generation model to synthesize $\mathbf{V}'$ from $(\mathbf{V}, \mathbf{G})$. We encode $\mathbf{V}$ and $\mathbf{V}'$ into video-VAE latents $\mathbf{z}=\mathrm{Enc}(\mathbf{V})$ and $\mathbf{z}'=\mathrm{Enc}(\mathbf{V}')$. We sample a timestep $\tau \in [0,1]$ and Gaussian noise $\boldsymbol{\epsilon}\sim\mathcal{N}(\mathbf{0},\mathbf{I})$, and construct the noised target latent along a straight-line path:
\begin{equation}
\mathbf{z}'_{\tau} = (1-\tau)\mathbf{z}' + \tau\boldsymbol{\epsilon}.
\end{equation}
The denoiser predicts a velocity field conditioned on the noised target latent, the source-video latent, and the geometry instruction:
\begin{equation}
\mathbf{v}_{\theta} = \mathbf{v}_{\theta}(\mathbf{z}'_{\tau}, \mathbf{z}, \mathbf{G}, \tau).
\end{equation}
We use the conditional flow-matching objective:
\begin{equation}
\mathcal{L}_{\mathrm{FM}} =
\mathbb{E}_{\mathbf{z}',\mathbf{z},\mathbf{G},\tau,\boldsymbol{\epsilon}}
\left[
\left\|
\mathbf{v}_{\theta}(\mathbf{z}'_{\tau},\mathbf{z},\mathbf{G},\tau)
-
(\boldsymbol{\epsilon}-\mathbf{z}')
\right\|_2^2
\right].
\end{equation}
Here $(\boldsymbol{\epsilon}-\mathbf{z}')$ is the target velocity induced by the straight-line rectified-flow path. Conditioning on $\mathbf{z}$ encourages preservation of input appearance and background, while $\mathbf{G}$ specifies the intended pre/post geometric state transition.

\begin{figure*}[t]
  \centering
  \includegraphics[width=\textwidth]{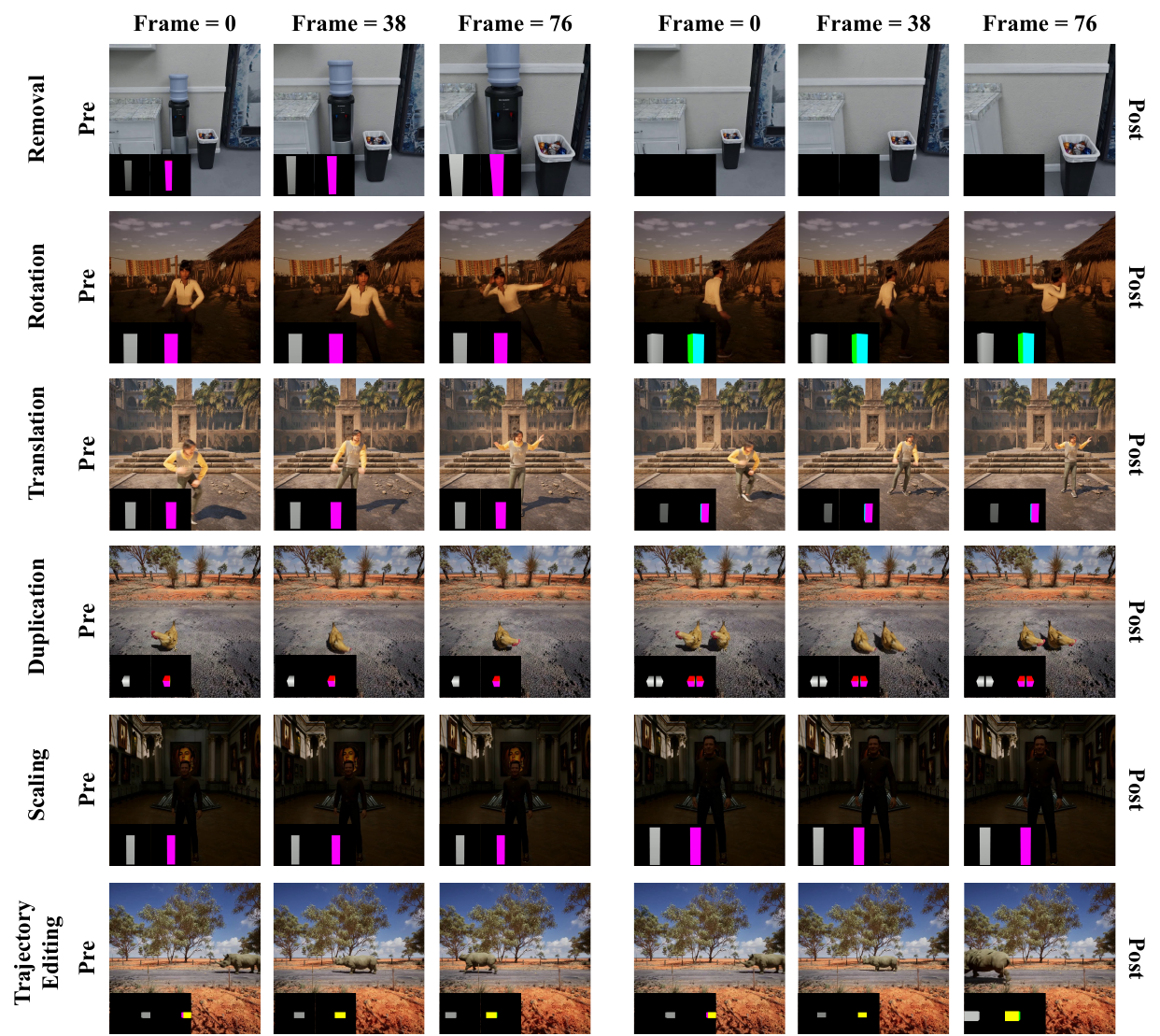}
  \vspace{-8pt}
  \caption{\textbf{\dataset\ examples across six geometric operators.}
  Each row corresponds to one operator.
  For each example, we render paired clips $(\mathbf{V}, \mathbf{V}')$ with the same scene, target identity, and camera trajectory; the post-edit clip differs only by the sampled operator and its parameters.
  We overlay the aligned geometry instruction streams as insets (depth-box and orientation-box visualizations) for the corresponding frames.}
  \label{fig:supp_givedata_examples}
  \vspace{-10pt}
\end{figure*}

\section{More Details on \dataset}
\label{sec:supp_givedata}

\paragraph{Overview.}
We construct \dataset\ using a fully automated rendering pipeline in Unreal Engine 5 by sampling \emph{edit programs} and rendering paired clips.
Concretely, we combine (i) diverse 3D scenes as backgrounds, (ii) target assets/instances placed in these scenes, (iii) motion clips for articulated subjects (when applicable), (iv) randomized camera trajectories, and (v) a representative set of DCC-style geometric operators with \emph{explicitly sampled parameters} (e.g., rotation angles, translation offsets, scale factors, duplication counts, and target trajectories).
Crucially, each training example in \dataset\ is a paired clip $(\mathbf{V}, \mathbf{V}')$ rendered with the \emph{same} scene, target identity, and camera trajectory; $\mathbf{V}'$ differs from $\mathbf{V}$ only by applying the sampled operator to the target while keeping all other factors fixed.
Along with the paired videos, we export per-frame geometry instruction streams aligned with the RGB frames.
Fig.~\ref{fig:supp_givedata_examples} visualizes representative examples for all six operators in \dataset, showing paired clips $(\mathbf{V}, \mathbf{V}')$ under identical scene/camera settings.
Each row corresponds to one operator, and we additionally overlay the aligned geometry instruction streams (depth/orientation box visualizations) as insets.

\paragraph{Scenes and target assets.}
We instantiate \dataset\ with $30$ distinct UE5 scenes spanning indoor/outdoor environments, and $319$ target instances from categories including humans, animals, vehicles, and rigid props.
All involved assets (scenes, models, and animations) are collected from the Fab marketplace.

\paragraph{Articulated motion and dynamic content.}
For articulated subjects (e.g., humans/animals), we attach motion clips during rendering.
We collect 92 motion clips and 20 facial-expression sets for human actors from the Fab marketplace.

\paragraph{Camera trajectories.}
We employ multiple camera-motion families to vary viewpoint while keeping paired supervision.
We follow ReCamMaster~\cite{bai2025recammaster} for trajectory families and sampling rules, except that we normalize the camera--target distance by target scale (instead of using a fixed absolute range) to accommodate objects of different sizes.

\paragraph{Operator sampling.}
We instantiate \dataset\ with DCC-style operators with explicit geometric targets, including rotation, translation, scaling, removal, duplication, and trajectory editing.
For each example, we sample operator parameters to form diverse edit programs:
(i) \textbf{Rotation}: rotations are predominantly yaw about the object up axis (Z), with angles sampled from $[30^\circ,180^\circ]$ with random sign; a small subset uses non-yaw axes for non-upright or wall-mounted objects.
(ii) \textbf{Translation}: 3D offsets are sampled proportional to the target 3D box scale (size-normalized), with constraints to avoid degenerate out-of-view cases.
(iii) \textbf{Scaling}: uniform 3D scaling factors are sampled from $[0.5,2]$, with the object kept grounded.
(iv) \textbf{Removal}: removal pairs are obtained by rendering an oracle clean plate with the target instance hidden, keeping all other scene elements and the camera identical.
(v) \textbf{Duplication}: we sample the number of copies from $2$ to $4$ and place each copy with size-normalized offsets and simple collision-aware checks.
(vi) \textbf{Trajectory editing}: we sample a target motion path with $1$ to $3$ keypoints and size-normalized displacement magnitudes, while keeping camera motion unchanged.
Operator frequencies are chosen to ensure sufficient coverage and stable training.

\paragraph{Scale and statistics.}
In total, \dataset\ contains 202k paired clips across 30 scenes and 319 target instances, rendered at 100 frames per clip and $1024\times1024$ resolution.
The dataset includes 56k rotation, 49k translation, 32k scaling, 45k removal, 5k duplication, and 15k trajectory-editing pairs.

\section{Additional Qualitative Comparisons}
\label{sec:supp_qual}

\begin{figure*}[t]
  \centering
  \includegraphics[width=\textwidth,height=0.9\textheight,keepaspectratio]{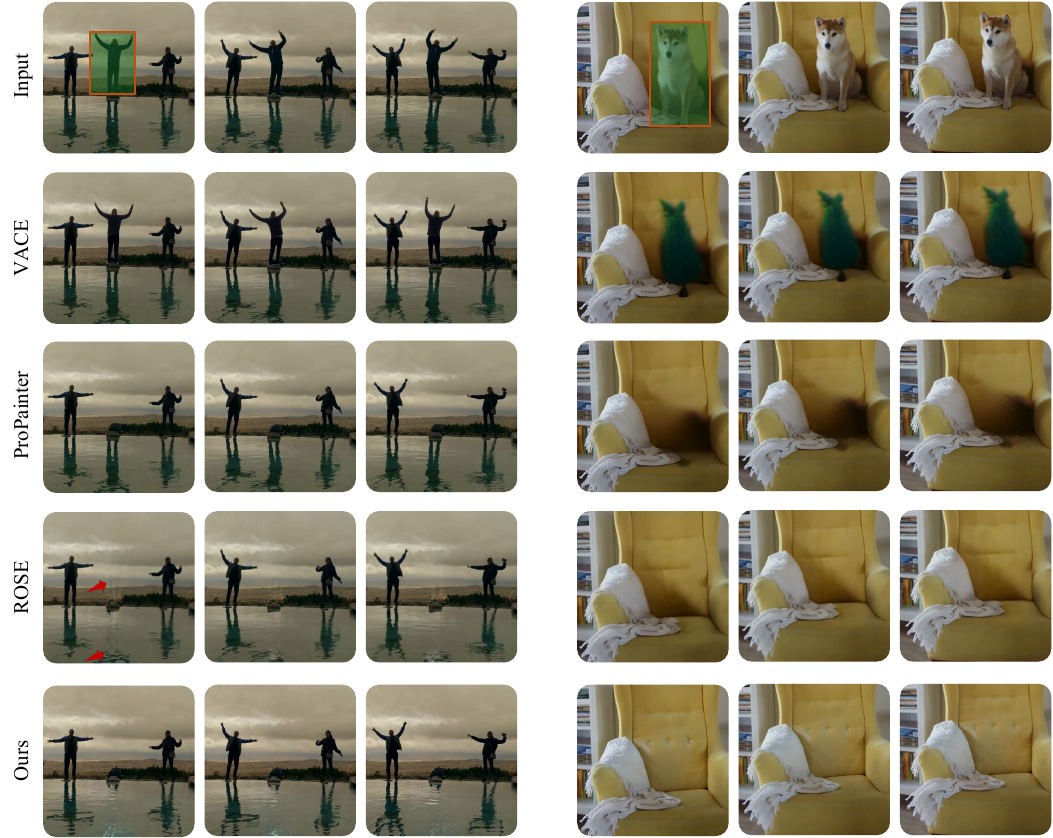}
  \vspace{-6pt}
  \caption{\textbf{Additional qualitative comparison: removal.}
We compare \method\ with video removal/inpainting baselines using the same target masks.
\method\ more reliably removes the target while preserving non-edited regions and coherently updating shadows and reflections.
Baselines may leave object remnants, retain inconsistent secondary effects, or introduce flickering artifacts.}
  \label{fig:supp_removal}
\end{figure*}

\begin{figure*}[t]
  \centering
  \includegraphics[width=\textwidth,height=0.9\textheight,keepaspectratio]{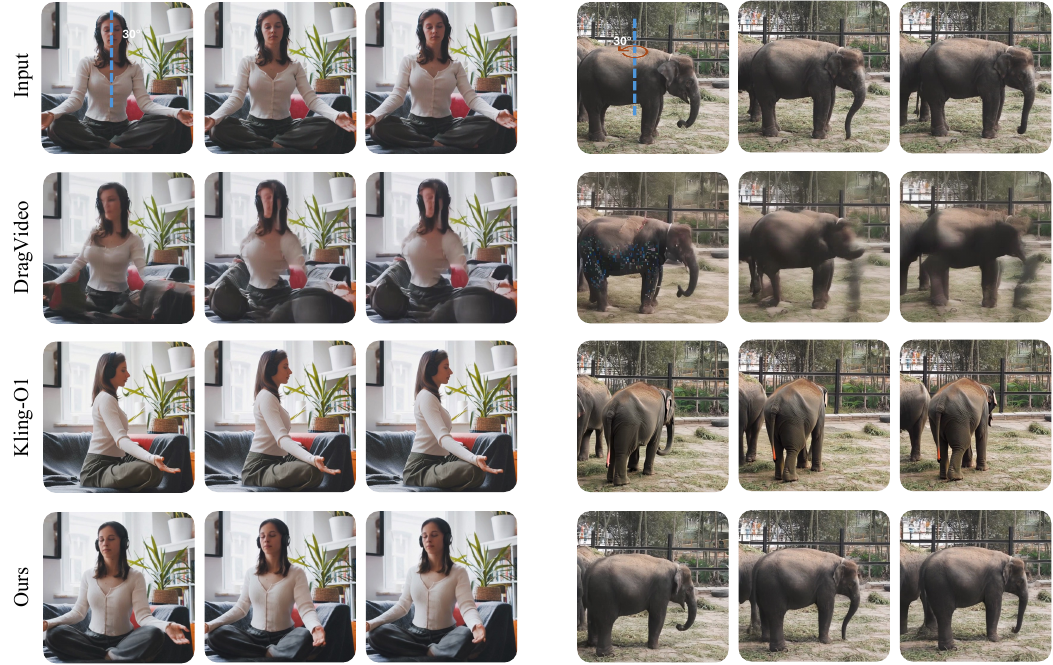}
  \vspace{-6pt}
  \caption{\textbf{Additional qualitative comparison: rotation.}
We compare rotation results on real videos under the same input and intended 3D heading change.
\method\ better matches the target orientation while preserving object identity, motion, and background consistency.
Baselines often under-rotate, rotate in the wrong direction or by an incorrect angle, or introduce identity and temporal drift.}
  \label{fig:supp_rotation}
\end{figure*}

\begin{figure*}[t]
  \centering
  \includegraphics[width=\textwidth,height=0.9\textheight,keepaspectratio]{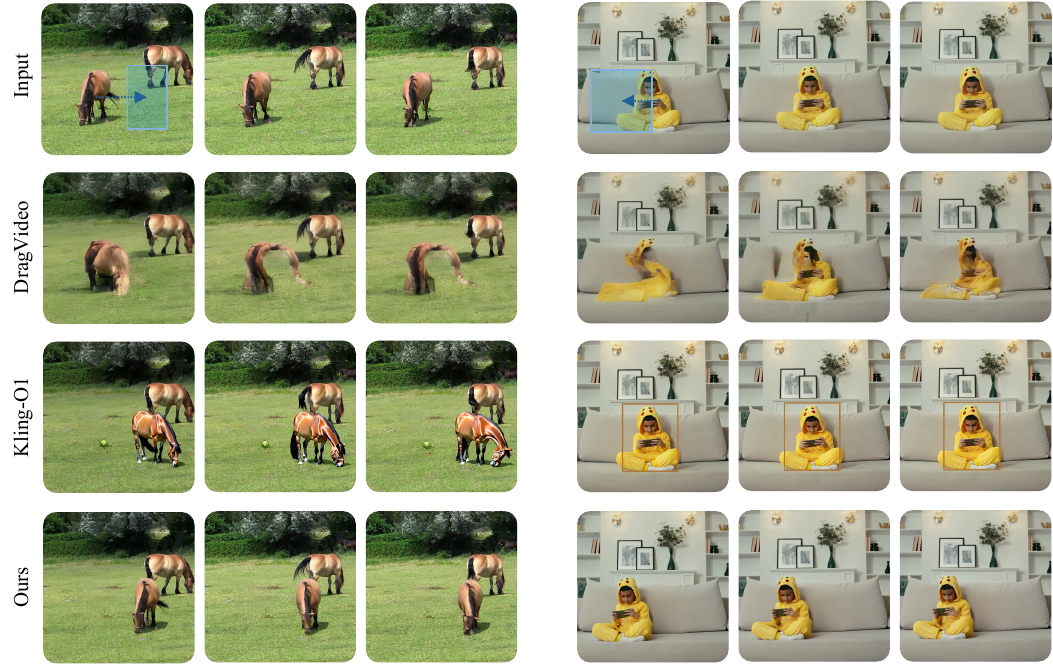}
  \vspace{-6pt}
  \caption{\textbf{Additional qualitative comparison: translation.}
We compare translation results on real videos under the same intended target displacement.
\method\ better matches the desired object location while preserving identity, motion, non-edited regions, and geometry-dependent effects.
Baselines often under-edit, drift from the target position, or introduce temporal inconsistency in the edited object.}
  \label{fig:supp_translation}
\end{figure*}

\begin{figure*}[t]
  \centering
  \includegraphics[width=\textwidth,height=0.9\textheight,keepaspectratio]{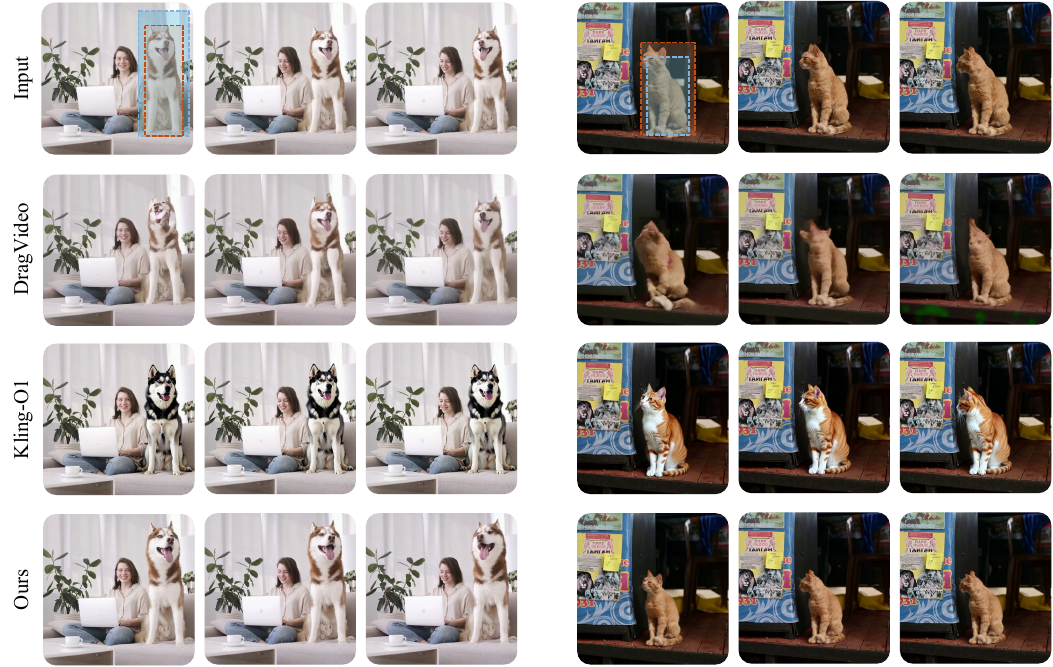}
  \vspace{-6pt}
  \caption{\textbf{Additional qualitative comparison: scaling.}
We compare scaling results on real videos under the same intended scale change.
\method\ changes the object scale more faithfully while preserving target identity, motion, and temporal coherence.
Baselines may under-scale or distort the object.}
  \label{fig:supp_scaling}
\end{figure*}

\begin{figure*}[t]
  \centering
  \includegraphics[width=\textwidth,height=0.9\textheight,keepaspectratio]{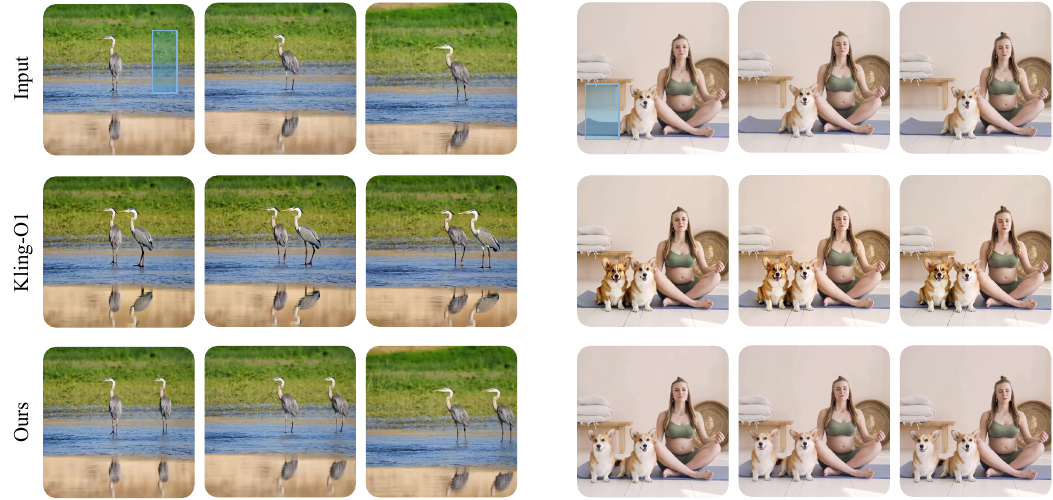}
  \vspace{-6pt}
  \caption{\textbf{Additional qualitative comparison: duplication.}
We compare duplication results on real videos under the same intended target-instance layout.
\method\ produces additional instances with more consistent identity, placement, and temporal coherence.
Baselines often fail to create all requested copies or introduce temporal inconsistency in the edited object.}
  \label{fig:supp_duplication}
\end{figure*}

\begin{figure*}[t]
  \centering
  \includegraphics[width=\textwidth,height=0.9\textheight,keepaspectratio]{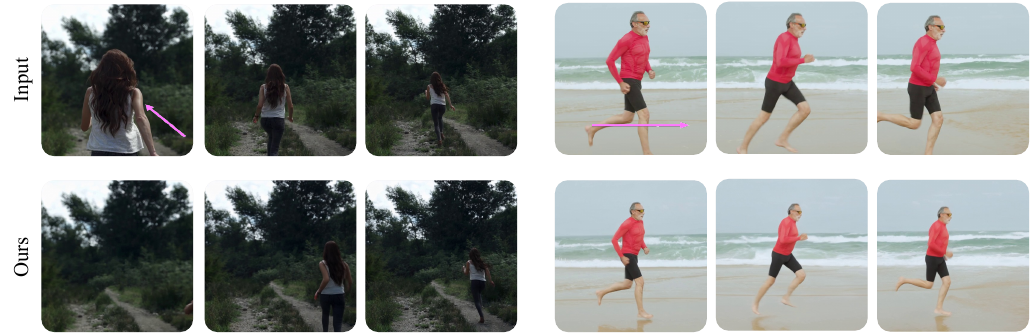}
  \vspace{-6pt}
  \caption{\textbf{Advanced trajectory editing with time-dependent targets.}
Our trajectory instruction specifies the target object's 3D state over time, rather than anchoring the edit to a modified first frame.
This supports delayed entrance, where the target is absent or outside the view in early frames and appears later, as well as trajectory retargeting, where an existing motion path is shifted to a different 3D location while preserving the original camera motion.}
  \label{fig:supp_traj_advanced}
  \vspace{-8pt}
\end{figure*}

\begin{figure*}[t]
  \centering
  \includegraphics[width=\textwidth,height=0.9\textheight,keepaspectratio]{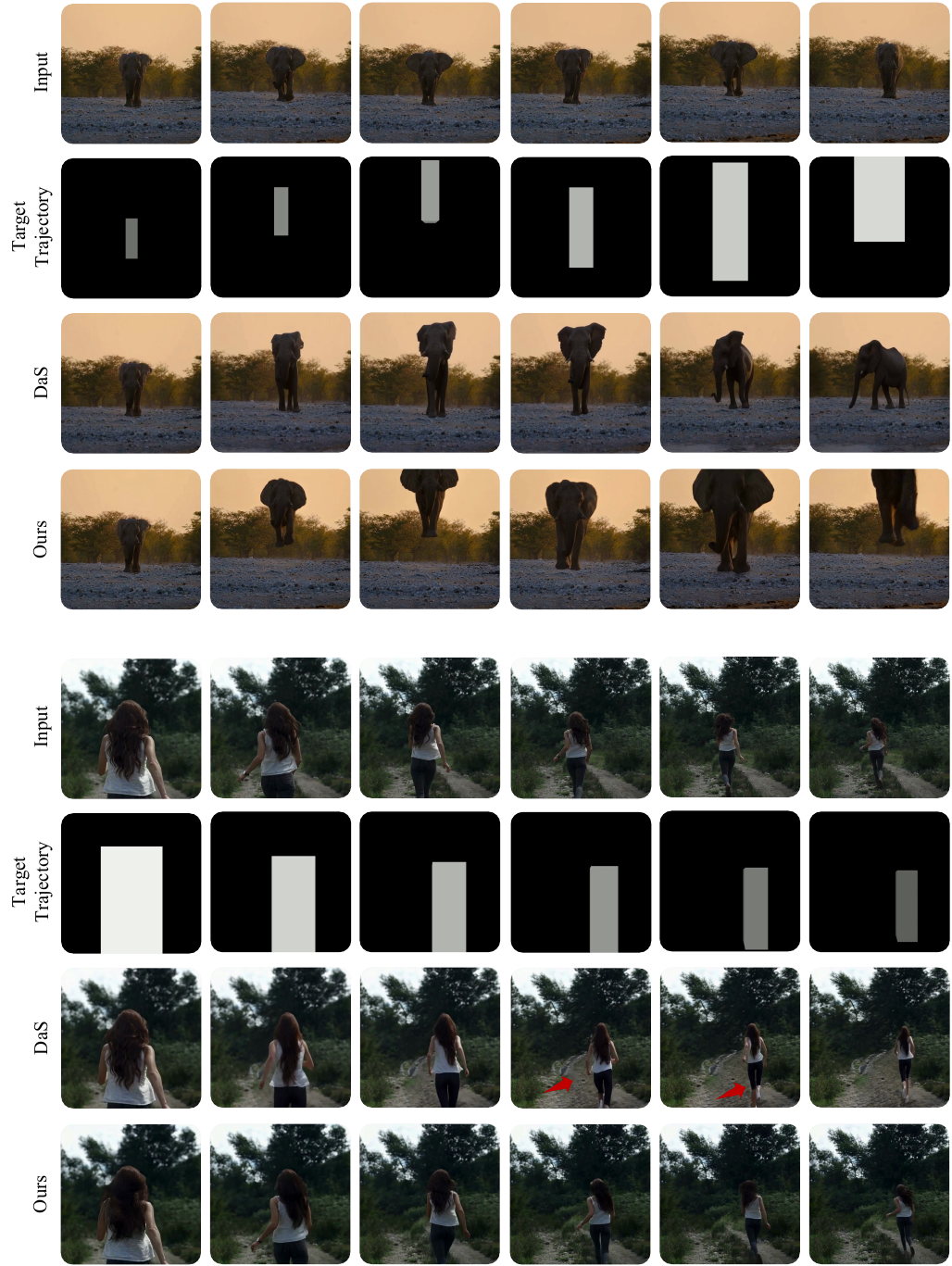}
  \vspace{-6pt}
  \caption{\textbf{Additional qualitative comparison: trajectory editing.}
We compare trajectory-editing results on real videos under the same intended motion path.
\method\ better follows the specified trajectory while preserving target identity and temporal coherence.
Baselines may deviate from the intended path or lose object consistency.
For visualization, we overlay the depth-box stream to illustrate the target trajectory.}
  \label{fig:supp_trajectory}
\end{figure*}

We provide operator-wise qualitative comparisons on real videos in Figs.~\ref{fig:supp_removal}--\ref{fig:supp_trajectory}, covering removal, rotation, translation, scaling, duplication, and trajectory editing.
Each figure shows representative frames from the input video, the editing instruction, and the outputs of \method\ and the corresponding baselines under the same case.
These comparisons highlight edit adherence, temporal consistency of the edited object, background preservation, and geometry-dependent secondary effects.

In addition to the representative frames shown in this document, we provide more video comparisons and qualitative results in the supplementary webpage included with the supplementary material. This webpage provides more side-by-side video comparisons and qualitative results across all operators, making it easier to examine temporal coherence, edit adherence, secondary effects update, and background preservation beyond static-frame visualizations.

Fig.~\ref{fig:supp_traj_advanced} further illustrates trajectory editing with time-dependent object-state targets.
Since our geometry instruction specifies the target object's 3D state over the video timeline, the edit need not be anchored to an edited first frame.
This enables both delayed entrance, where the target appears only in later frames, and trajectory retargeting, where an existing motion path is shifted to a different 3D location, such as farther from the camera.

\section{Additional Analyses}
\label{sec:supp_analyses}

\subsection{Robustness to estimated geometry}
The controlled benchmark uses engine-rendered geometry instructions, while real videos rely on estimated geometry from the SAM~2 + Geo4D pipeline. 
To measure the effect of this gap, we replace the ground-truth pre-edit geometry on \benchmark\ with estimator-derived geometry from the same real-video acquisition pipeline, while keeping the target edit unchanged. 
The cross-operator average changes from $22.70/0.8006/0.1439$ to $22.32/0.7968/0.1471$ in PSNR/SSIM/LPIPS, indicating only modest degradation under realistic geometry-estimation noise.

\subsection{Conditioning Efficiency}
\label{sec:supp_conditioning_efficiency}

Recent diffusion-transformer video models have increasingly explored sequence-level conditioning to integrate heterogeneous signals, such as text, camera, layout, or motion cues~\cite{ju2025fulldit,ye2025unic}.
Inspired by this design principle, we represent the source video and the pre-/post-edit geometry instructions as separate token streams.
Token-level conditioning preserves the identity of the source, pre-edit, and post-edit streams, which improves edit faithfulness but increases transformer sequence length compared with channel-wise fusion.
Tab.~\ref{tab:conditioning_efficiency} analyzes this quality--efficiency tradeoff under different fusion strategies and GI temporal strides.
For token-level conditioning, the geometry-instruction streams are temporally downsampled by a stride $s$ before tokenization, reducing the sequence-length increase compared with using full-frame instruction streams.
We measure latency under the same resolution, frame length, sampler, number of denoising steps, precision, and hardware; latency refers to the denoising stage and excludes video decoding, geometry estimation, and VAE decoding.

At the same GI stride $s=2$, token concatenation improves the cross-operator average from $19.82$ to $22.70$ in PSNR, from $0.7073$ to $0.8006$ in SSIM, and from $0.2393$ to $0.1439$ in LPIPS, while increasing denoising latency from $94$s to $120$s.
Using denser GI streams with $s=1$ slightly improves quality but substantially increases latency to $198$s.
Increasing the stride to $s=4$ reduces latency to $97$s, at the cost of lower edit faithfulness than the default $s=2$ setting, but it still outperforms channel-wise fusion.
We therefore use token-level conditioning with GI stride $s=2$ as the default implementation, balancing edit quality and inference efficiency.
For efficiency-sensitive scenarios, a larger GI stride provides a practical lower-cost alternative by reducing the number of GI tokens.

\begin{table}[t]
\caption{\textbf{Conditioning efficiency analysis.}
We compare fusion strategy and GI temporal stride. 
Channel and token fusion are compared at the same GI stride $s=2$.
Token rows with different $s$ analyze the quality--efficiency tradeoff of temporal GI subsampling. $^\dagger$ denotes the default setting.}
\label{tab:conditioning_efficiency}
\vspace{-4pt}
\centering
\small
\setlength{\tabcolsep}{3.0pt}
\renewcommand{\arraystretch}{1.0}
\begin{tabular}{lccccc}
\toprule
\thead{Fusion} &
\thead{GI stride $s$} &
\thead{PSNR$\uparrow$} &
\thead{SSIM$\uparrow$} &
\thead{LPIPS$\downarrow$} &
\thead{Latency$\downarrow$} \\
\midrule
Channel concat & 2 & 19.82 & 0.7073 & 0.2393 & 94s \\
Token concat & 1 & 22.97 & 0.8186 & 0.1402 & 198s \\
Token concat$^\dagger$ & 2 & 22.70 & 0.8006 & 0.1439 & 120s \\
Token concat & 4 & 21.92 & 0.7896 & 0.1645 & 97s \\
\bottomrule
\end{tabular}
\vspace{-6pt}
\end{table}

\begin{figure}[t]
  \centering
  \includegraphics[width=\columnwidth]{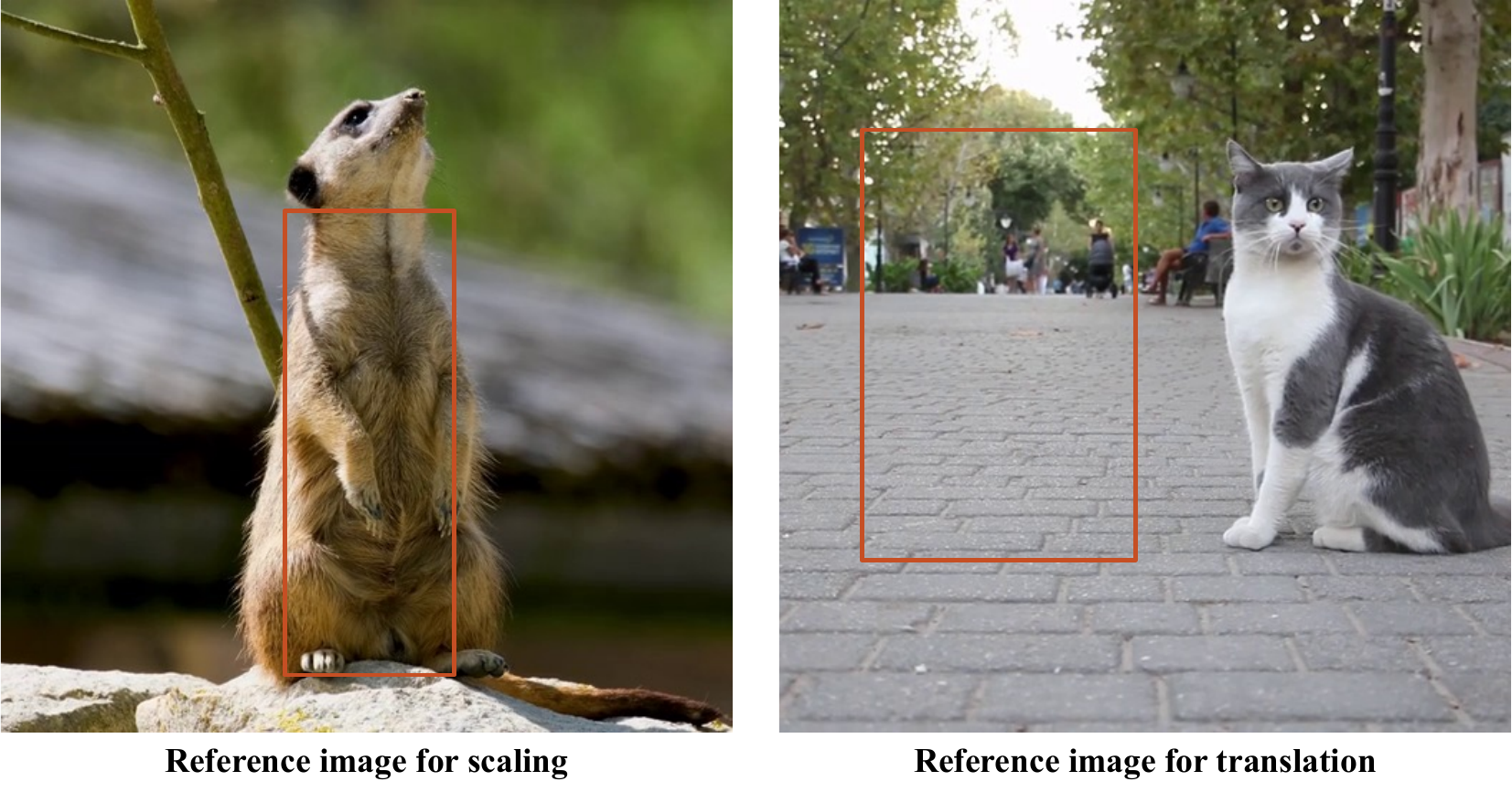}
  \vspace{-12pt}
  \caption{\textbf{Kling-O1 reference image for visual-signal-guided editing.}
  We overlay 2D box(es) to indicate the desired post-edit state for translation/scaling/duplication following the instructions in~\cite{team2025kling}.}
  \label{fig:kling_refimg}
  \vspace{-8pt}
\end{figure}

\begin{figure}[t]
  \centering
  \includegraphics[width=\columnwidth]{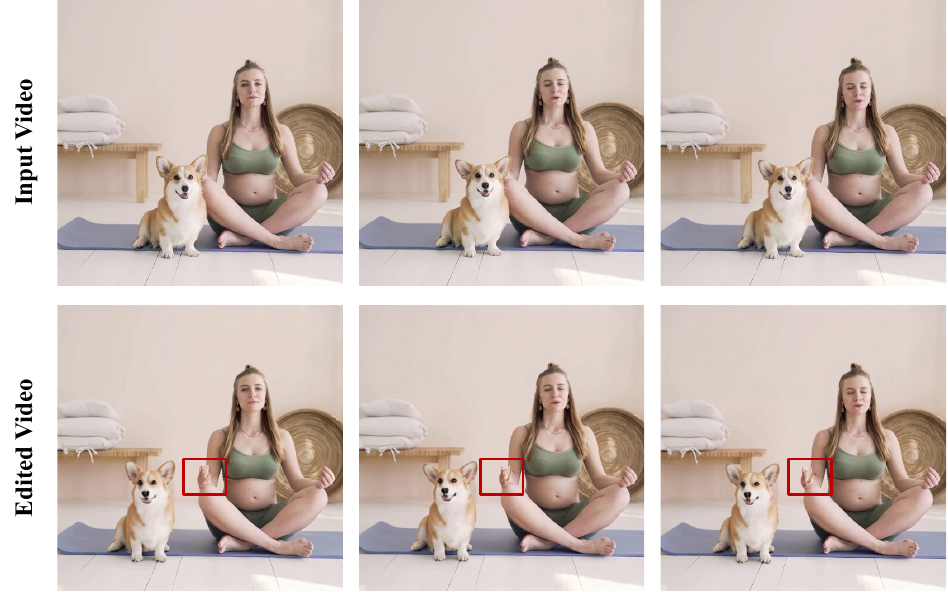}
  \vspace{-6pt}
  \caption{Failure case visualization.}
  \label{fig:supp_failures}
  \vspace{-8pt}
\end{figure}

\section{More Experimental Details}
\label{sec:supp_baselines}

\paragraph{\dataset\ Curation.}
We train \method\ on \dataset, a paired-video dataset rendered by UE5 with aligned geometry-instruction streams for object-level edits. Each training example contains an input video $\mathbf{V}$, an engine-rendered target $\mathbf{V}'$, and four instruction streams $\{\mathbf{V}^{d}_{\mathrm{pre}}, \mathbf{V}^{o}_{\mathrm{pre}}, \mathbf{V}^{d}_{\mathrm{post}}, \mathbf{V}^{o}_{\mathrm{post}}\}$. 
All of the rendered video clips are standardized to a resolution of $1024 \times 1024$ and 100 frames.

\paragraph{\benchmark\ Curation.}
Existing video editing evaluations rarely provide paired references with explicit geometric targets.
We therefore introduce \benchmark, a curated benchmark of paired edits rendered by UE5 under controlled settings.
Each benchmark sample provides an input video $\mathbf{V}$ and its geometry-instruction streams,
and uses the engine-rendered edited video $\mathbf{V}'$ as the reference.
\benchmark\ covers $6$ operators including rotation, removal, scaling, translation, duplication, and trajectory editing, with $30$ clips per operator and $180$ clips in total. The benchmark spans diverse scenes, assets, and camera-motion families. To ensure fair evaluation, we select \benchmark\ such that both the scene and the target asset instance are disjoint from \dataset\ used for training.

\paragraph{Real-Video Set Curation.}
To assess transfer to real-world videos, we collect $240$ in-the-wild clips from~\cite{Pexels} covering diverse categories and camera motions. We filter for clips with a clearly identifiable target object visible for most of the clip and without hard scene cuts. For evaluation, geometry instructions are obtained from lightweight user specification, combined with SAM 2 segmentation and Geo4D geometry estimation; when automatic estimation is noisy, we allow minimal user refinement of the box to ensure a well-posed instruction.

\paragraph{Evaluation Metrics.}
We report metrics on two complementary evaluation settings.
(i) \textbf{Reference-based metrics on \benchmark.}
Since \benchmark\ provides paired ground-truth references $\mathbf{V}'$, we evaluate reconstruction and perceptual similarity using PSNR, SSIM, and LPIPS~\cite{ssim,zhang2018unreasonable} between generated outputs $\hat{\mathbf{V}}$ and $\mathbf{V}'$. Since baselines may output different numbers of frames and the compared method set varies by operator, we perform per-operator temporal alignment by mapping all results for operator $\ell$ to a common $T_\ell$-frame timeline via temporal subsampling and compute metrics on the aligned frames.
(ii) \textbf{Reference-free metrics and human evaluation on real videos.}
On the real-video set, we report three aggregated VBench scores (Temporal, Consistency, and Quality) for overall video quality.
We further conduct a best-of-$K_\ell$ user study, here $K_\ell$ equals the number of compared methods for operator $\ell$. Given the input video and the editing instruction, participants are shown the edited results from multiple methods for the same case and select a single best result.
Participants are instructed to jointly consider (a) whether the result follows the editing instruction, (b) whether the edited object remains temporally consistent (identity and motion) without drift, (c) whether non-edited regions (background and other objects) are preserved, and (d) whether geometry-dependent secondary effects (e.g., shadows and reflections) remain coherent and plausible after the edit.
Method identities are hidden and the presentation order is randomized per trial; participants may replay videos before deciding.
In total, we recruit $30$ participants and collect 7200 trials.
For each operator $\ell$, we report \emph{WinRate} (\%) as the fraction of trials in which a method is selected as the single best result among the $K_\ell$ candidates shown for that operator.
Since the compared method set differs across operators, we report WinRate per operator rather than aggregating across operators.

\paragraph{Baseline selection rationale.}
\method\ is evaluated in a video-input editing setting, where the goal is to edit a target object in a source clip according to a specified object-level transformation, while preserving non-edited content over time. We therefore prioritize baselines that natively operate on source videos, including general-purpose video editors and operator-specific methods with the closest task overlap and executable control interface. For these primary baselines, we provide the same source clip and translate the same edit intent into the method's native input modality, such as prompts, masks, drag handles, point tracks, or reference images. Kling-O1~\cite{team2025kling} is included as a commercial general-purpose editor to compare our explicit 3D-aware instruction interface against mainstream prompt- or visual-signal-guided editing. DragVideo~\cite{deng2024dragvideo} provides drag-based control for video editing, making it a relevant baseline for geometric changes in our video-input setting, while image-conditioned drag-based generation methods are discussed in the related work.
For removal, we include ProPainter~\cite{zhou2023propainter} and ROSE~\cite{miao2025rose}, which are operator-specific video inpainting/removal methods. We also include VACE~\cite{jiang2025vace}, a unified video generation and editing model; among its official task interfaces, video inpainting/removal is the most directly aligned with our benchmark scope, so we evaluate VACE on the removal operator.
For trajectory editing, we additionally include DaS~\cite{gu2025diffusion} as a closest-interface reference baseline. DaS is not designed for direct source-video editing; instead, it performs controllable video generation from 3D track-based guidance. Nevertheless, its control interface is closely related to \method's geometry instructions, since both represent object motion using explicit 3D spatial guidance over time.

\paragraph{Per-baseline inputs.}
We describe how we instantiate the inputs for each baseline in our experiments, and how we translate our geometry instructions into their supported control modalities. Unless otherwise specified, all methods take the same input clip as introduced in the main paper and we use the official implementations / interfaces with recommended default settings.

\paragraph{Common preprocessing.}
For all baselines, we resize frames to $512\times512$, and use the same $T$ frames per clip for the corresponding operator (benchmarks and real videos follow the protocol introduced in the main paper), and apply the same per-operator temporal alignment when computing PSNR/SSIM/LPIPS on \benchmark.

\textbf{(i) DragVideo~\cite{deng2024dragvideo}.}
DragVideo is controlled by sparse handle/target points~\cite{deng2024dragvideo}.
To make the comparison reproducible, we deterministically convert our pre-/post-edit 3D box states into sparse 2D point constraints:
(1) we sample a fixed set of candidate 3D anchor points on the 3D box,
(2) project these anchors to image space using the camera model (oracle on \benchmark; estimated on real videos) and retain only anchors whose projections lie inside the target-object mask, to avoid background points when the box is looser than the object,
(3) treat each retained projection in the pre-state as a handle and its corresponding projection in the post-state as the target.
We follow the official DragVideo implementation and default settings.

\textbf{(ii) Kling-O1~\cite{team2025kling}.}
Kling-O1 is a closed-source, multimodal commercial video generation model that supports visual-signal-guided editing.
We therefore report the exact interface we used.
Across all tasks, we keep the input clip fixed and use a short, operator-specific prompt template that explicitly enforces the key invariances required by our evaluation: \emph{the camera motion should remain unchanged}, the target object should remain \emph{temporally consistent} (identity preserved and motion consistent with the input video), and the background should be preserved as much as possible, while allowing only physically plausible secondary-effect updates.
For \textit{translation, scaling, and duplication}, text-only instructions can be ambiguous in direction, magnitude, and the intended post-edit state.
Following the Kling technical report interface~\cite{team2025kling}, we provide an additional \emph{single reference image} per case to disambiguate the target post-edit state.
Specifically, we overlay one or multiple 2D bounding boxes derived from our 3D box instructions on the \textbf{first frame} to indicate the desired post-edit state: one box for translation/scaling, and one or multiple boxes for duplication (one box per instance).
An example of this reference-image visual signal is shown in Fig.~\ref{fig:kling_refimg}.
In all these tasks, Kling-O1 is prompted to match the post-edit box specification in the reference image while keeping camera/background and target motion consistent with the original clip (Tab.~\ref{tab:supp_kling_inputs}).
For \textit{rotation}, we do not provide a reference image and instead use a concise text prompt that specifies a rotation by an angle around the object's up axis.
Since our rotation evaluation is restricted to yaw rotations, we specify clockwise (or counterclockwise) rotation as viewed from above, and apply the same invariance constraints as above (Tab.~\ref{tab:supp_kling_inputs}).
We use the same prompt template across cases within each operator, filling only the operator parameters.

\begin{table}[t]
\caption{\textbf{Kling-O1 input instantiation per operator.}
We report the exact interface we used. For translation/scaling/duplication, we additionally provide a first-frame reference image with 2D box overlays derived from projected pre/post 3D boxes to reduce ambiguity.}
\label{tab:supp_kling_inputs}
\vspace{-4pt}
\centering
\scriptsize
\setlength{\tabcolsep}{2.2pt}
\renewcommand{\arraystretch}{1.10}

\begin{tabularx}{\columnwidth}{l p{0.24\columnwidth} X}
\toprule
\textbf{Task} & \textbf{Inputs} & \textbf{Text prompt template} \\
\midrule

Rot. &
Video + Text &
Rotate the target object by <angle> clockwise (or counterclockwise) around its own up axis as viewed from above. Keep the camera motion unchanged. Keep the target's motion the same as in the input video. Preserve the target's identity and keep the background unchanged. Maintain plausible shadows and reflections. \\

\addlinespace[2pt]
Trans. &
Video + RefImg + Text &
Move the target object to match the box shown in the reference image. Keep the camera motion unchanged. Keep the target's motion the same as in the input video. Preserve the target's identity and keep the background unchanged. Maintain plausible shadows and reflections. \\

\addlinespace[2pt]
Scale &
Video + RefImg + Text &
Scale the target object to match the box shown in the reference image. Keep the camera motion unchanged. Keep the target's motion the same as in the input video. Preserve the target's identity/appearance and keep the background unchanged. Maintain plausible shadows and reflections. \\

\addlinespace[2pt]
Dupl. &
Video + RefImg + Text &
Duplicate the target object to match the boxes shown in the reference image with one box per instance. Keep the camera motion unchanged. Keep the motion consistent with the input video for all instances. Preserve identity for all copies and keep the background unchanged. Maintain plausible occlusions, shadows, and reflections. \\

\bottomrule
\end{tabularx}

\vspace{-9pt}
\end{table}

\textbf{(iii) ProPainter~\cite{zhou2023propainter} and ROSE~\cite{miao2025rose}.}
These methods address video inpainting / object removal and take a video and a binary mask as input~\cite{zhou2023propainter,miao2025rose}.
For fair comparison, we use the same target masks across methods, obtained from SAM 2.
We run the official implementations with recommended settings.

\textbf{(iv) VACE~\cite{jiang2025vace}.}
VACE is a unified video generation and editing model with multiple supported task interfaces.
Among its official settings, the interface most directly matched to our benchmark scope is video inpainting/removal.
We therefore evaluate VACE on the removal operator.
For each case, we use the same input video and target mask as the other removal baselines, and run VACE in its official inpainting/removal setting with recommended parameters.

\textbf{(v) DaS~\cite{gu2025diffusion} for trajectory editing.}
DaS is an I2V diffusion model conditioned on a 3D tracking video~\cite{gu2025diffusion}.
On \benchmark, since DaS cannot take the full source video as input, we use the first frame of the reference edited video $\mathbf{V}'$ as an oracle starting image (an optimistic setting for DaS) and generate the tracking video from our geometry instructions following DaS.
On real videos, a privileged edited reference frame is unavailable. To enable a comparable I2V evaluation, we restrict trajectory edits to preserve the initial object state (i.e., the trajectory starts from the original position), so that the input video's first frame can be used as condition image for DaS. We then derive the 3D tracking video from the estimated geometry instructions using the same sampling and encoding rule, and compare DaS and \method\ under this matched-start protocol.
Because DaS is conditioned on a single image and a tracking video rather than the full source clip, it does not observe later-frame appearance, articulation, or motion cues of the original target, which can make identity and motion preservation more challenging in our source-video editing setting.

\paragraph{On reconstruction-heavy 3D proxy baselines.}
Reconstruction-based method such as Shape-for-Motion~\cite{liu2025shape} operate in a different test-time regime: they reconstruct a per-video 3D proxy and then apply edits with additional optimization before rendering/refinement.
While such explicit structure can improve edit adherence when reconstruction is reliable, the approach incurs substantial per-video overhead and is sensitive to reconstruction quality.
For example, Shape-for-Motion reports $\sim$90 minutes for reconstruction of a video on an A100 GPU, plus additional mesh editing time ranging from $\sim$30 seconds to $\sim$10 minutes depending on the editing type, and $\sim$43 seconds for generative rendering~\cite{liu2025shape}.
Given this regime mismatch and cost, we do not include such pipelines in our large-scale quantitative evaluation.
Instead, we provide a representative failure case in the main paper and discuss when reconstruction brittleness affects downstream editing.

\paragraph{Realistic Paired Removal Benchmark}
\label{sec:supp_realistic_removal}

To complement the engine-rendered paired evaluation on \benchmark, we construct a realistic paired removal benchmark using $30$ real videos from Pexels.
Following the copy-paste protocol of ROSE~\cite{miao2025rose}, we select videos where an object-free background region can be used to synthesize an approximate clean reference after removing the target object.
All methods use the same input video and target mask, and generated results are evaluated against the constructed reference using PSNR, SSIM, and LPIPS on temporally aligned frames.
As shown in Tab.~\ref{tab:supp_realistic_removal}, \method\ achieves the best PSNR/SSIM/LPIPS among the compared removal methods, supporting the trend observed on the engine-rendered paired benchmark.

\begin{table}[t]
\caption{\textbf{Realistic paired removal benchmark.}
We evaluate removal results on 30 Pexels videos with approximate clean references constructed following the copy-paste protocol. 
All methods use the same input videos and target masks.}
\label{tab:supp_realistic_removal}
\vspace{-4pt}
\centering
\small
\setlength{\tabcolsep}{4pt}
\renewcommand{\arraystretch}{1.0}
\begin{tabular}{lccc}
\toprule
Method & PSNR$\uparrow$ & SSIM$\uparrow$ & LPIPS$\downarrow$ \\
\midrule
ProPainter & 27.21 & 0.8794 & 0.1403 \\
VACE & 25.38 & 0.8514 & 0.1536 \\
ROSE & 27.83 & 0.8819 & 0.1478 \\
\method\ (Ours) & \textbf{28.41} & \textbf{0.8875} & \textbf{0.1296} \\
\bottomrule
\end{tabular}
\vspace{-6pt}
\end{table}

\paragraph{Public-Backbone Instantiation}
\label{sec:supp_wan}

To assess portability, we instantiate \method\ on the public Wan2.1~\cite{wan2025video} backbone while keeping the same pre/post depth-box and orientation-box instruction format and conditional training objective.
We include this experiment as an implementation check on an open backbone, rather than as a replacement for the main evaluation.
On \benchmark, the Wan2.1 instantiation achieves an average PSNR/SSIM/LPIPS of $21.87/0.7907/0.1658$ across the six operators.
This experiment supports the portability of the proposed pre/post object-state representation beyond our main backbone.

\section{Failure Cases and Limitations}
\label{sec:supp_failures}

We visualize a representative failure case in Fig.~\ref{fig:supp_failures}.
As \method\ builds on a text-to-video diffusion backbone, it can inherit limitations in synthesizing fine-grained details.
In this example, translating the dog reveals a previously occluded region (the woman’s hand).
While \method\ achieves the intended translation and preserves the overall scene, the newly exposed moving hand exhibits inferior quality as shown in the second row.
Additionally, as mentioned in the main paper, concatenating multiple geometry-instruction streams with video tokens improves generation quality but increases computational cost.
% This reflects a common challenge for occlusion-changing edits: reliably completing unseen content, especially thin articulated parts, remains difficult.

\paragraph{Societal considerations.}
Like other generative video editing systems, \method\ could be misused to alter object placement, presence, or motion in realistic footage.
We encourage responsible use, clear disclosure of edited content, and appropriate safeguards for sensitive media.

% \clearpage

% Bibliography
\bibliographystyle{ACM-Reference-Format}
\bibliography{paper}

\end{document}